\documentclass[11pt]{article}

\usepackage[margin=1in]{geometry}
\usepackage{authblk}
\usepackage{graphicx}
\usepackage{amsmath}
\usepackage{amssymb}
\usepackage{caption}
\usepackage{setspace}
\usepackage[numbers]{natbib}
\usepackage{subcaption}

\usepackage{times}
\usepackage{soul}
\usepackage{url}
\usepackage[hidelinks]{hyperref}
\usepackage[utf8]{inputenc}
\usepackage{caption}
\usepackage{amsmath}
\usepackage{amsthm}
\usepackage{booktabs}
\usepackage{algorithm}
\usepackage{algorithmic}
\usepackage[switch]{lineno}
\usepackage{tikz}
\usetikzlibrary{automata, positioning, arrows.meta, calc}
\usepackage{subcaption}
\usepackage{enumitem}

\title{A Machine Learning Framework for Off Ball Defensive Role and Performance Evaluation in Football}

\author[1,2]{Sean Groom}
\author[2]{Francisco Belo}
\author[2]{Axl Rice}
\author[3]{Liam Anderson}
\author[1*]{Shuo Wang}
\affil[1]{School of Computer Science, The University of Birmingham, Birmingham, UK}
\affil[2]{Nottingham Forest Football Club, Nottingham, UK}
\affil[3]{School of Sport, Exercise and Rehabilitation Sciences, The University of Birmingham, Birmingham, UK}
\affil[*]{Corresponding author: s.wang.2@bham.ac.uk}

\date{}

\begin{document}

\maketitle

\begin{abstract}
\noindent Evaluating off-ball defensive performance in football is challenging, as traditional metrics do not capture the nuanced coordinated movements that limit opponent action selection and success probabilities. Although widely used possession value models excel at appraising on-ball actions, their application to defense remains limited. Existing counterfactual methods, such as ghosting models, help extend these analyses but often rely on simulating "average" behavior that lacks tactical context. To address this, we introduce a covariate-dependent Hidden Markov Model (CDHMM) tailored to corner kicks, a highly structured aspect of football games. Our label-free model infers time-resolved man-marking and zonal assignments directly from player tracking data. We leverage these assignments to propose a novel framework for defensive credit attribution and a role-conditioned ghosting method for counterfactual analysis of off-ball defensive performance. We show how these contributions provide a interpretable evaluation of defensive contributions against context-aware baselines.

\end{abstract}

\section*{Introduction}

Elite sporting organizations are increasingly leveraging machine learning to inform tactical planning and player evaluation. This advancement is fueled by rich datasets, including event data (e.g., passes and shots) and player tracking data from multi-camera systems \cite{Spatio-Temporal-Analysis-of-Team-Sports, match-analysis-big-data-and-tactics-current-trends-in-elite-soccer}. For instance, pitch control models \cite{RN31, RN35, RN30} estimate a team's probability of controlling different areas of the field. While these models are valuable for visualizing territorial dominance, they often fail to appraise the contextual value of those areas within a specific game state. Similarly, possession value models \cite{Rudd2011MarkovSoccer, Singh2018xT, Cervone2016EPV, RN22, RN21}, which appraise on-ball actions, face significant challenges when applied to defensive performance analysis \cite{RN4}. This is because defensive play presents unique appraisal problems, player's individual actions are not the only important consideration, but also their coordinated movements with teammates, which makes it difficult to isolate individual contributions. Furthermore, successful defensive play is often defined by actions that don't occur, i.e. preventing a pass or a shot \cite{Tuyls2020GamePWB, Merhej2021WhatHNA, Forcher2023TheSFB}. Traditional metrics like tackles and interceptions are insufficient as they don't capture the subtle ways defenders constrain attacking options. Consequently, the analysis of defensive performance significantly lags behind that of offensive play.

To address the limitations of traditional metrics, recent research has explored counterfactual analysis as a means of evaluating defensive performance. A notable example is the use of "ghosting" models, which generate simulated player trajectories to analyze defensive behavior by measuring how a player’s actual movement alters offensive outcomes—such as pass completion or shot quality—when compared to a hypothetical "ghost" player \cite{Le2017DataDrivenGUC, Yurko2024NFLGAA, Seidl0BhostgustersRD}. This approach provides a powerful way to quantify a defender's impact beyond simple event counts. However, the application of these techniques in football remains limited \cite{RN90}, and they often rely on simulating "average" behavior, which may not align with a team's specific tactical system. More recent research has attempted to move beyond this by introducing latent role variables into generative or predictive models of player motion \cite{Scofano_2024, Fassmeyer2025Interactive}. While these roles help the model represent coordinated team structures, they are designed to capture statistical regularities in movement rather than interpretable tactical categories familiar to coaches (e.g., man-marking or zonal defending). To date, such models have been primarily evaluated for predictive accuracy, not for counterfactual assessment of tactical performance. Therefore, conditioning counterfactual generation on interpretable tactical roles may offer more relevant insights for coaching and performance analysis.

Attempts to value off-ball defence in football have historically been limited, due in part to the difficulty of linking defensive positioning to observable outcomes. Recently, graph-based deep learning approaches have begun to address this gap in open play. Everett et al. \cite{everett2025gapp} introduce GAPP, an attention-based framework that estimates Defender Influence by predicting pass reception probabilities using Graph Attention Networks. By leveraging attention weights and node-masking interventions, their method provides local, instance-level explanations of which defenders most strongly affect specific attacking options.

However, as established in prior work on model interpretability, attention mechanisms—while often predictive and locally faithful—do not in themselves constitute explanations of causal or semantic structure \cite{jain2019attention, wiegreffe2019attention}. In particular, attention weights are not guaranteed to correspond to uniquely meaningful roles or responsibilities unless such concepts are explicitly represented in the model. In the context of off-ball defending, this means that high attention or influence scores indicate that a defender affected an outcome, but not why this occured. Moreover, attention-masking-based influence measures implicitly compare observed behaviour to the absence of a defender’s contribution, rather than to a realistic alternative execution of the same defensive role, limiting their ability to support structurally grounded tactical reasoning.

In parallel, Kim et al. propose DEFCON \cite{kim2025defcon}, a probabilistic framework that assigns defensive credit by marginalising over learned action selection, success, and defender responsibility probabilities to estimate marginal reductions in Expected Possession Value (EPV). This formulation yields a principled and stable attribution of defensive value and avoids reliance on explicit attention masking, making it well-suited to aggregating player contributions across matches and contexts. Importantly, DEFCON evaluates defenders relative to the observed defensive organisation, attributing credit within a fixed tactical configuration.

Set pieces, particularly corner kicks, offer a structured and valuable context for studying defensive behavior. Occurring roughly ten times per match and accounting for a significant percentage of goals (11\% of all goals in the 2024/2025 English Premier League season, calculated from Statsbomb event data), corners are highly scripted, meaning that improvements in performance can lead to substantial gains without needing to acquire new personnel. Their impact is significant enough that many clubs now employ dedicated set-piece coaches. Defensive instructions during corners typically involve either tracking a specific opponent (man-marking) or zonal marking, where players guard predefined locations. Most teams adopt hybrid strategies that reflect their tactical identity. Prior work has largely focused on scheme identification and pattern discovery. For example, Shaw \& Gopaladesikan mapped recurring corner routines and inferred defender roles \cite{routineInspection}. Bauer et al. \cite{Bauer} developed an approach using a convolutional–recurrent neural networks (CNN–LSTM) to assign defenders various roles, including zonal defending and man-marking roles, yielding who was marking whom but at the cost of a hand-labelled dataset that is neither publicly nor commercially available. DeepMind and Liverpool FC's TacticAI utilises D\textsubscript{2}-equivariant graph neural networks to predict reception and shot outcomes, as well as generating suggestions on how to improve these outcomes \cite{RN8}. Although TacticAI's suggestions were preferred by expert raters, they are not explicitly parameterized by interpretable tactical roles. Accordingly, we seek a framework that infers time-resolved tactical roles from tracking data without time-intensive labeling. This framework will be coupled with outcome prediction models to facilitate counterfactual evaluation and individual defensive performance evaluation.

Addressing this gap, we introduce a covariate-dependent Hidden Markov Model (CDHMM) \cite{Bishop,RN45} specifically tailored for the analysis of corner kicks. In our model, the latent states are designed to represent interpretable tactical roles, such as which defender is assigned to man-mark a specific opponent and which defenders occupy zonal positions. The model's covariate-dependent transitions capture the dynamic nature of set-play behavior, allowing it to infer time-resolved tactical assignments and derive team- and delivery-specific zonal structures directly from tracking data without the need for manual labeling of roles. 

We leverage these role assignments in two key ways. First, they provide coach-facing profiles of defender and attacker behaviour at set plays and extend existing outcome models (e.g., reception probability, expected threat, shot likelihood) to the off-ball domain by attributing defensive credit to specific players in specific roles. Second, we define role-conditioned ghosts, counterfactual baselines that replace a defender with an average counterpart from the same tactical role and team context, enabling situation-specific, role-aware off-ball performance analysis. We illustrate practical utility via a reception-prediction case study and discuss generalization to other outcome models such as expected threat. This research was conducted in collaboration with, and validated by, professional football analysts and coaches from an elite level professional club, including the club's set-piece coach.

Our work is complementary in spirit but distinct in emphasis to Everett et al. and Kim et al. Rather than treating defensive organisation as implicit context, we model it explicitly through latent defensive roles, such as man-marking and zonal coverage, and their temporal evolution during set pieces. This representation enables both player-level evaluation and structurally grounded counterfactual analysis, allowing alternative defensive organisations or role assignments to be compared directly. As a result, while recent approaches quantify how much defensive value is realised within an existing structure, our framework makes it additionally possible to ask how defensive value would change under different, tactically plausible organisational choices, supporting decision-making and tactical analysis.

\textbf{Contributions}
\begin{itemize}
\item We introduce a label-free CDHMM for corner kicks that jointly infers time-resolved man-marking and zonal assignments, as well as team- and delivery-specific zonal structures, directly from tracking data. Building on our earlier IT4PSS workshop model \cite{62f0eb3414504f6dae4f8e5f6621bd43}, we extend the transition dynamics with a covariate-dependent formulation and provide substantially expanded validation and analysis.
\item We propose coach-facing behavioural metrics, inspired by the basketball HMM of Franks et al.~\cite{RN2}, derived from the inferred man-marking and zonal roles to provide interpretable insights for tactical analysis.

\item We present a novel method for off-ball defensive credit attribution. Unlike general open-play models that distribute credit via implicit responsibility scores \cite{kim2025defcon}, our method attributes value based on explicitly inferred, time-resolved tactical assignments.

\end{itemize}

To our knowledge, we present the first ghosting framework in football that conditions defender baselines on interpretable tactical roles learned directly from data. This enables a more structured and context-aware counterfactual evaluation of defensive performance and offers a distinct advantage over attention-masking techniques \cite{everett2025gapp} by enabling counterfactual comparisons against average role-specific behavior rather than the absence of a defender.

\section*{Results}

\subsection*{Data and Preprocessing}

This study utilizes tracking data covering four complete English Premier League seasons, from 2020/21 through to 2023/24. Corner kick sequences were extracted by aligning tracking data with synchronized event logs, resulting in an initial dataset of 14,678 corners. Corners involving red cards or substitution errors were excluded from analysis to ensure data quality. Individual CDHMMs were trained separately for each defending team, and further distinguished by corner kick delivery type, specifically categorizing sequences as either inswinging or outswinging. For downstream tasks, such as recipient and shot prediction described later in this work, short corners—defined as short passes to a nearby teammate rather than deliveries into the penalty area—were excluded based on event data annotations, resulting in a dataset of 13,752 corners. To standardize the data, all corner sequences were canonicalised: each sequence was translated and reflected so that all deliveries appear to originate from the top-right corner of the pitch, with the coordinate system's origin at the defending team's penalty spot. Sequences were captured starting one second before the corner kick and were truncated either two seconds after delivery or upon the second subsequent on-ball event, whichever occurred first. This approach aligns with a prior methodology \cite{routineInspection}. Finally, to simplify our analysis of defensive marking, goalkeepers were excluded from man-marking state considerations.

\subsection*{Inferred Defensive Behaviors from the CDHMM}

Our framework utilizes a covariate-dependent Hidden Markov Model (CDHMM) to analyze defensive behavior during corner kicks. The model represents each defender's actions as a sequence of latent states over time, which correspond to interpretable tactical roles, including man-marking a specific attacker or defending a fixed zonal region of the pitch. At each timestep, the model determines a defender's most likely state based on their observed position, and transitions between states are governed by probabilities that dynamically adapt to sequence specific covariates. This approach allows us to unsupervisedly infer tactical dynamics and role changes directly from tracking data, without the need for manual labels.

\begin{figure}[htbp]
    \centering
    \includegraphics[width=0.8\linewidth]{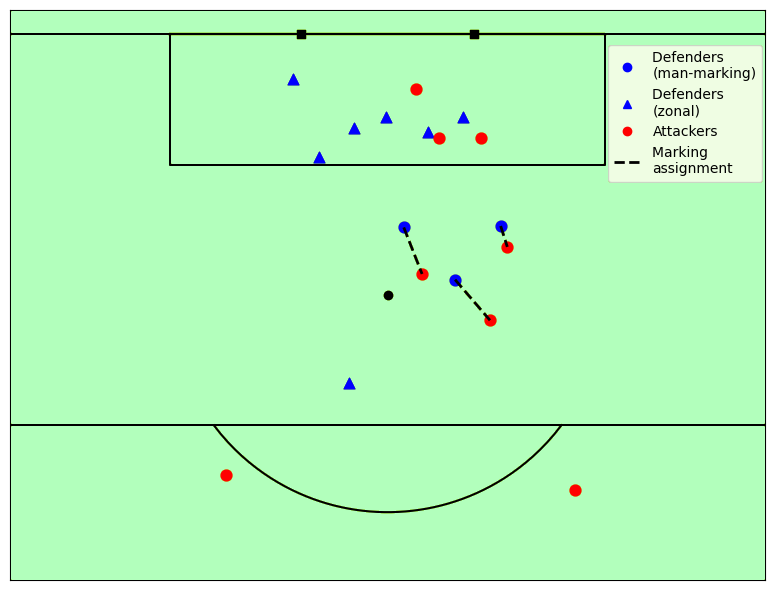}
    \caption{
        A visualization of a team defending an inswinging corner. Blue circles represent defenders assigned to man-marking roles, with dashed black lines connecting them to their assigned attacker (red circles). Blue triangles indicate defenders in zonal marking roles, positioned in a compact structure near the goalmouth.
    }
    \label{fig:man_marking_example}
\end{figure}

The CDHMM framework provides a granular, time-resolved analysis of defensive behavior. Figure \ref{fig:man_marking_example} visualizes the inferred defensive assignments at a single time step during a corner routine. The model distinguishes between man-marking roles (blue circles linked by dashed lines to an assigned attacker) and zonal defending roles (blue triangles positioned in a compact structure across the goalmouth). This mixture of roles accurately reflects the hybrid defensive strategies common in professional football. These roles and assignments are not pre-defined; instead, they emerge directly from the model’s unsupervised learning of spatio-temporal patterns, highlighting its ability to capture complex tactical structures without manual labeling.

The CDHMM's emission distribution defines the likelihood of a defender's position given their inferred tactical role. For man-marking states, this distribution depends on the assigned attacker's location, modeling how closely a defender tracks their opponent. For zonal states, emissions are modeled as bivariate Gaussians centered on fixed regions of the pitch. A key feature of our approach is that these zonal emission distributions are learned separately for each team and corner delivery type (inswinging or outswinging), enabling the model to capture the nuanced spatial structures of different defensive schemes.

\begin{figure*}
   \centering
   \begin{subfigure}[b]{0.5\textwidth}
      \centering
      \includegraphics[width=\textwidth]{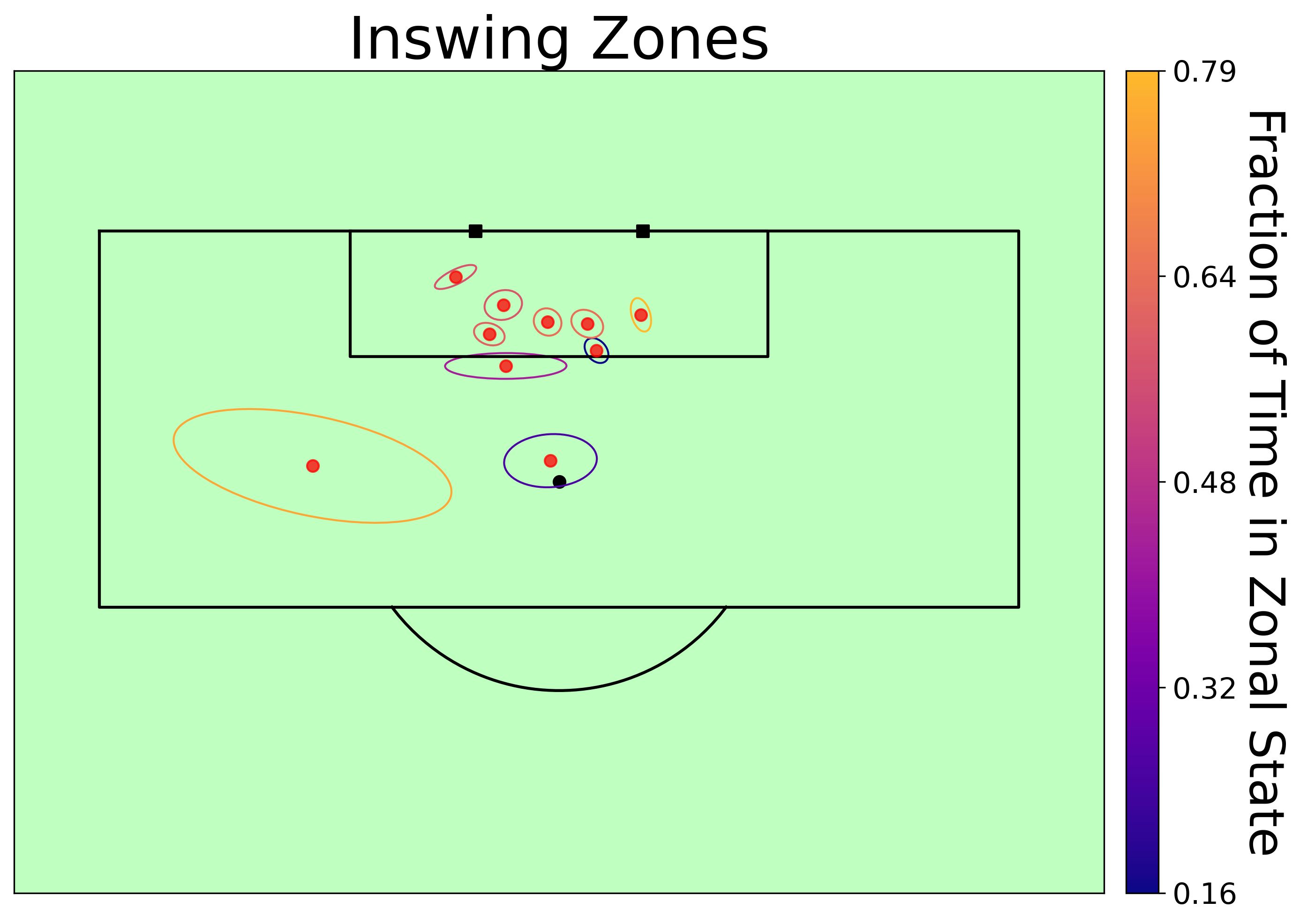}
      \caption{Estimated zonal states for a team defending
       \newline inswinging corners.}
      \label{fig:inswinging}
   \end{subfigure}%
   \hfill
   \begin{subfigure}[b]{0.5\textwidth}
      \centering
      \includegraphics[width=\textwidth]{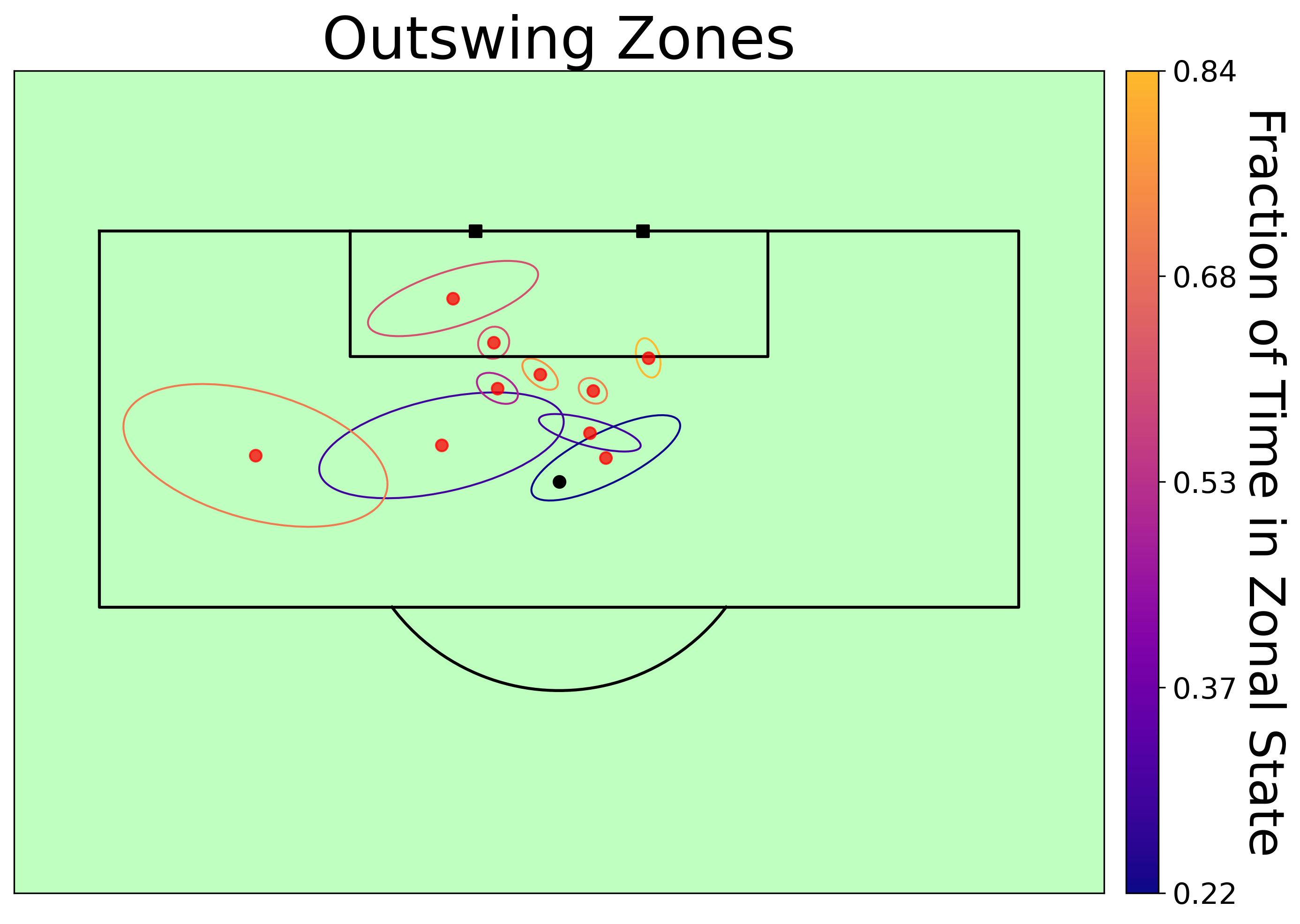}
      \caption{Estimated zonal states for a team defending \newline
      outswinging corners.}
      \label{fig:outswinging}
   \end{subfigure}
   \caption{This figure illustrates the estimated zonal states for a team from the 2023/24 Premier League season when defending against inswinging and outswinging corner deliveries. Each sub-figure presents 10 zonal emission distributions, marked by a red circle indicating the mean position and encircled by a 66\% confidence interval. The colour of each ellipse denotes the average fraction of the corner sequence duration that a defender spends in that zonal state before transitioning to a man-marking state. Pitch markings are shown in black; goalposts are represented by black squares and the penalty spot by a black circle.}
   \label{fig:MU_team_zones}
\end{figure*}

Figure \ref{fig:MU_team_zones} visualizes the estimated zonal distributions for a single team during the 2023/24 season. Each zone is represented by a bivariate Gaussian, with the red dot indicating the estimated mean position and the ellipse showing the 66\% confidence interval. The color of the ellipse encodes the average duration a defender spends in that zone before transitioning to a man-marking assignment.

The results reveal that zonal structures differ markedly between delivery types. For inswinging corners, zones tend to concentrate near the goalmouth, while for outswinging corners, the structure is more dispersed, with coverage extending toward the edge of the penalty area. This highlights the model's sensitivity to tactical context.

In addition to learning zonal structures, the CDHMM also estimates how defenders position themselves relative to their assigned attacker when engaged in man-marking. A key parameter governing this behaviour is $\gamma_o$, that governs how tightly defenders position themselves relative to their assigned attacker. This parameter captures the extent to which a defender "shades" toward the attacker along the line connecting the attacker to the goalmouth, which can be interpreted as marking tightness.

To account for spatial variation, we estimate separate $\gamma_o$ values for each $3\,\text{m} \times 3\,\text{m}$ bin on the pitch, allowing the model to flexibly capture how man-marking behaviour changes across different areas. In general, we find that defenders mark more tightly as attackers approach the goal, reflecting the natural compression of player spacing and increased defensive urgency in central areas.

\begin{figure}[htbp]
    \centering
    \includegraphics[width=\linewidth, trim=0 130 0 0, clip]{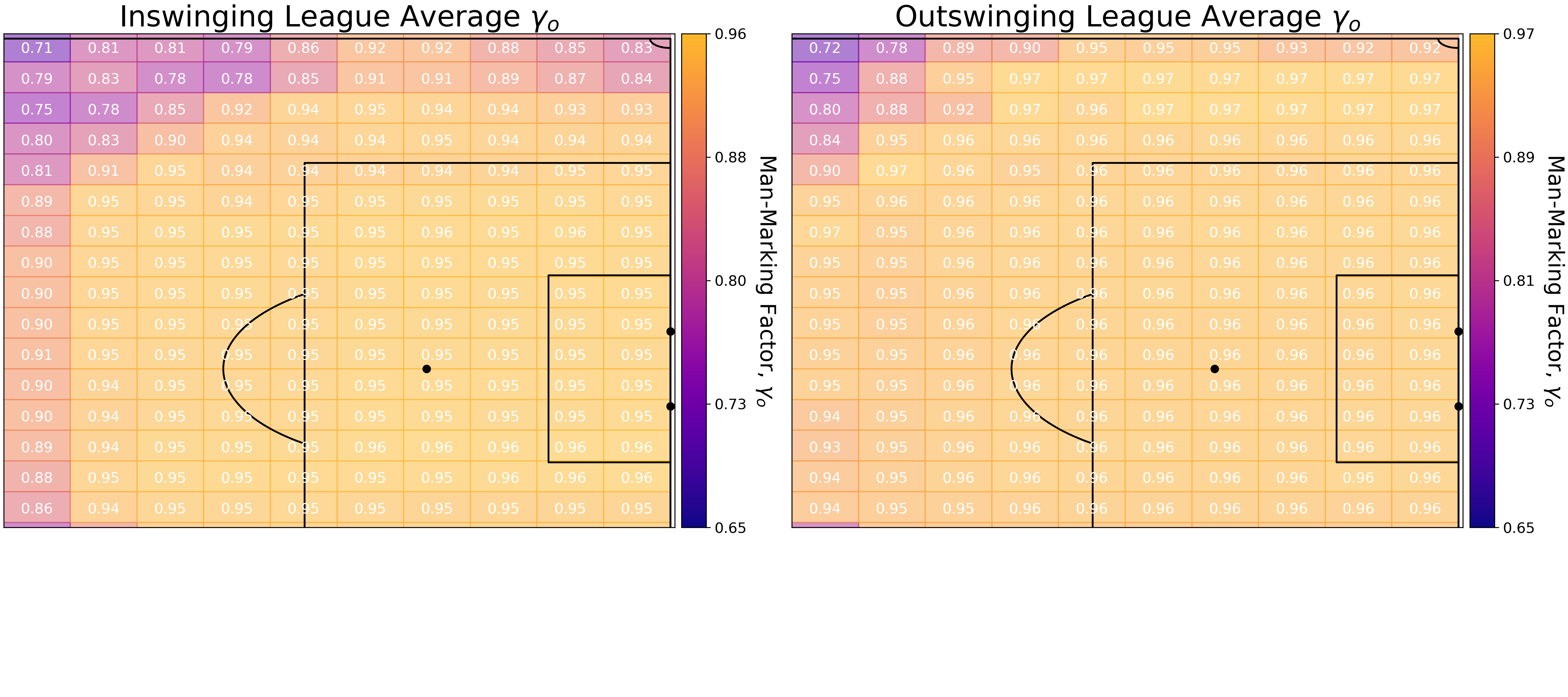}
    \caption{
        A comparison of league-average man-marking factors, $\gamma_o$, across the pitch for inswinging (left) and outswinging (right) corner deliveries. The heatmaps reveal differences in defensive behaviour, outswinging corners are associated with slightly tighter man-marking across a broader area, particularly beyond the near-post zone. In both cases, the average offset between defenders and their assigned attackers decreases as the attacker nears the goal.
    }
    \label{fig:gamma heatmap}
\end{figure}

Figure~\ref{fig:gamma heatmap} shows 2023/24 league-average $\gamma_o$ values across the pitch for inswinging and outswinging deliveries. These maps are produced by averaging the learned parameters from each team's CDHMM across an entire season. The results reveal subtle but consistent differences in defensive behaviour across delivery types: outswinging corners are associated with slightly tighter marking over a broader region, particularly beyond the near-post zone.

\subsubsection*{Covariate-Dependent Transition Model}

A core feature of the CDHMM is its covariate-dependent transition model, which allows transition probabilities between states to vary dynamically with the game state. The model learns linear weight vectors ($\beta$) that govern different types of transitions. For example, $\beta_m$ determines the probability of a defender continuing to mark the same attacker, while $\beta_z$ controls the probability of a defender remaining in a zonal role. We also define a switching parameter, $\beta_s$, which governs the probability of a defender changing man-marking assignments or transitioning from a zonal role to a man-marking one.  Transitions from man-marking to zonal states are not permitted in our model based on discussions with coaches and analysts.  This dynamic transition structure enables the model to capture the fluid nature of defensive roles during a set piece.  A full list of covariates and details of how these probabilities are computed are provided in the Methods section.

\begin{figure}[htbp]
    \centering
    \includegraphics[width=\linewidth]{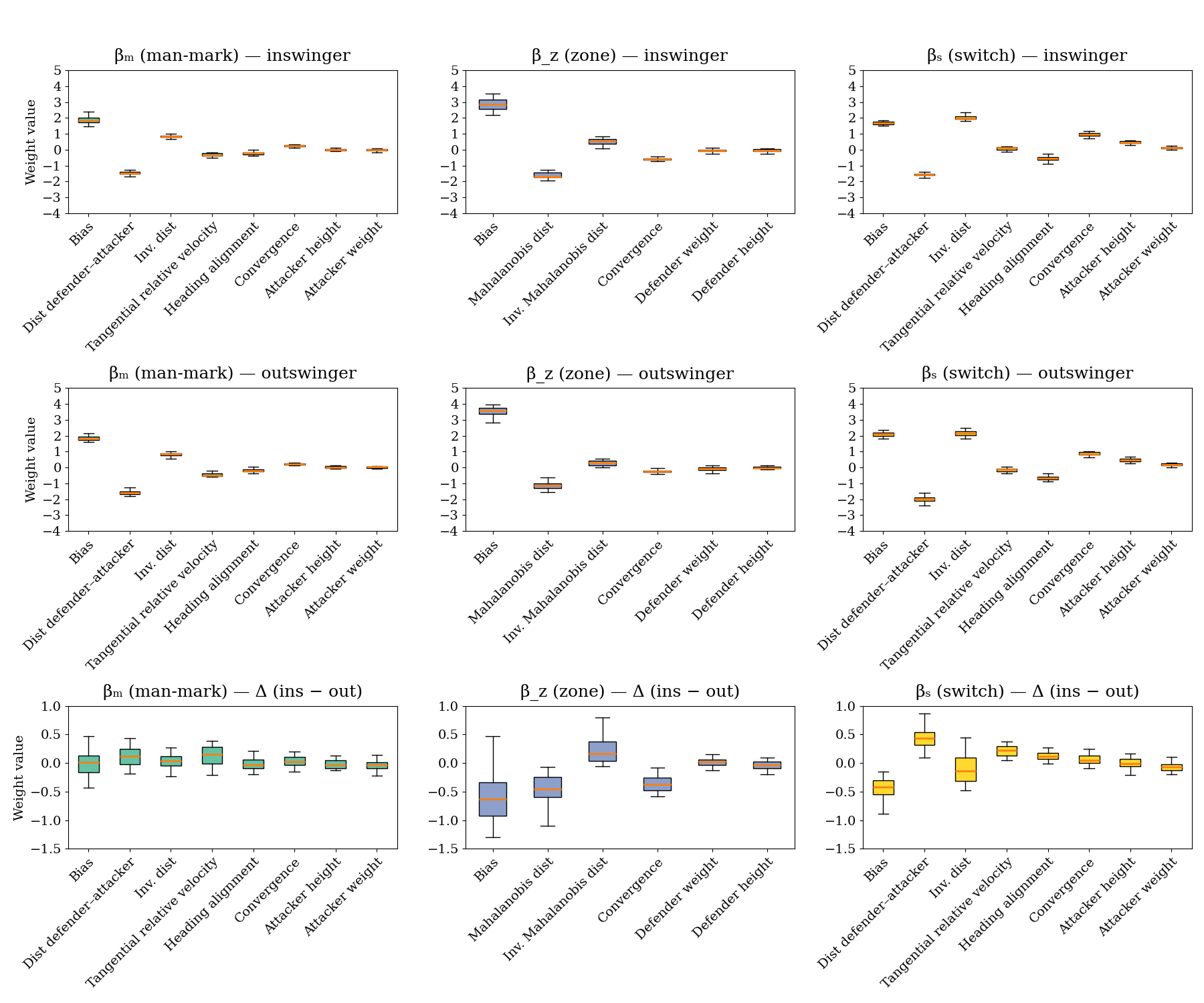}
    \caption{Distribution of learned transition model parameters for different defensive behaviours during all 2023/24 season corner kicks. Each column corresponds to a transition type: $\beta_m$ for continuing to man-mark the same attacker, $\beta_z$ for continuing to zonally defend a designated zone, and $\beta_s$ for switching from zonal or one man-marking state to another man-marking assignment. The top two rows show the distribution of parameter weights across all teams for inswinging and outswinging deliveries, respectively. The bottom row illustrates within-team differences between inswinging and outswinging deliveries, highlighting delivery-specific tactical adaptations. These parameters are estimated from team and delivery specific Hidden Markov Models.}
    \label{fig:beta box plots}
\end{figure}

The CDHMM's dynamic transition model provides insights into how defensive behaviors evolve during a corner. Figure \ref{fig:beta box plots} shows the distribution of learned transition weight parameters ($\beta$) across all teams and delivery types.

We find that some aspects of transition behaviour are consistent across teams. For example, the relatively narrow and approximately zero-centered, distribution of $\beta_m$ values for inswinging corners (top left) suggests that teams adopt broadly similar policies for maintaining man-marking assignments. The bottom left panel further confirms this, showing only minor within-team changes in man-marking behaviour between delivery types.

By contrast, more variation is observed in zonal self-transitions. The distribution of $\beta_z$ parameters (top middle) is wider, and the corresponding bottom middle plot reveals large within-team differences between inswinging and outswinging corners. This reflects the influence of delivery-specific zonal configurations on defensive behaviour — particularly the fact that different zones are occupied in each delivery type and that defenders tend to remain in zonal roles for different durations depending on the delivery.

\begin{figure}[htbp]
    \centering
    \includegraphics[width=\linewidth]{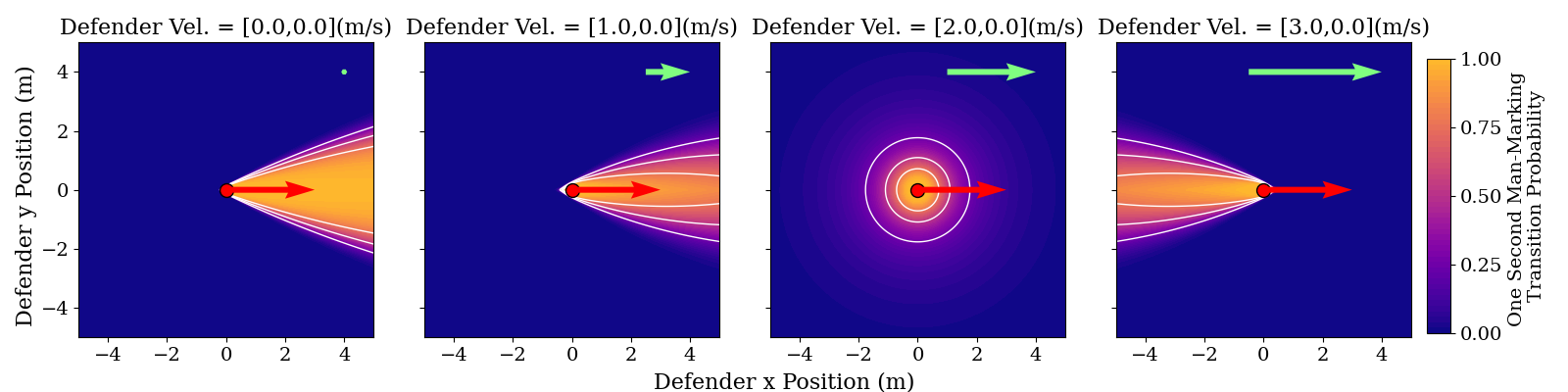}
    \caption{
        Each subplot shows how the probability of continuing to man-mark a specific attacker ($p_m$) varies spatially under different defender velocities for an individual team in the 2023/24 season during inswinging corners. The attacker is fixed at the origin $(0, 0)$ and moves rightward with velocity $(2.0, 0.0)$, with height $1.85\,\text{m}$ and weight $82\,\text{kg}$. In all panels, the defender moves in the same direction (rightward), with speed increasing from left to right. Red arrows indicate the attacker's velocity vector; green arrows in the top-right corner of each plot indicate the defender’s velocity. We plot three iso-probability contours at $p_m = 0.25$, $0.5$, and $0.75$ to highlight decision boundaries. As defender speed increases, the model assigns lower $p_m$ values to trailing positions and higher values to regions more aligned with the attacker's trajectory, reflecting the influence of velocity alignment on predicted marking persistence. Note that since data is sampled at 25 frames per second, the one-second transition probability of maintaining a man-marking assignment is given by $p_m^{25}$.
        }

    \label{fig:short transition heatmap}
\end{figure}

The CDHMM's dynamic transition model provides further insights into how defensive behaviors evolve based on player movement. Figure \ref{fig:short transition heatmap} visualizes the one-second man-marking transition probability, $p_m^{25}$ derived from a single team's $\beta_m$ parameters. In these heatmaps, the attacker is fixed at the origin and moves to the right, while the defender's velocity increases across the panels from left to right. The heatmaps show the spatial distribution of the continuation probability, revealing how the model dynamically adapts to the relative motion between the players. As the defender's speed increases, the model assigns lower probabilities to trailing positions and higher values to regions more aligned with the attacker's trajectory. This demonstrates the model's ability to capture the influence of velocity alignment on the predicted persistence of a man-marking assignment. Additional examples of learned transition surfaces under different configurations are provided in the Appendix (Figure ~\ref{fig:transition heatmaps}), further illustrating the flexibility of the model in capturing context-dependent marking behaviours.

\subsubsection*{Latent state–derived characterisation of player behaviour}

The learned parameters and outputs of the CDHMM provide detailed, time-resolved insights into team defensive behavior during corner sequences, including team-specific zonal configurations and per-frame marking assignments. These outputs are leveraged to derive a set of player-level metrics that characterize individual roles and behaviors within corner routines.

\subsubsection*{Context-Aware Man-Marking Attention}

In their work in basketball analysis Franks et al.\cite{RN2} introduce man-marking attention and defensive entropy, defining man-marking attention, $A_{k}^{(p)}$, as a metric associated with each attacking player $k$ that is calculated as the total number of frames the attacker $k$ is marked divided by the sequence length $T_p$, while defensive entropy captures how frequently teams assign multiple defenders to a single attacker. In a similar approach, we adapt the man-marking attention metric to account for the possibility of zonal defending during the corner sequences. 

To account for the possibility that defenders may be zonally marking, we define the context-aware man-marking attention score, $\mathrm{CA}_k$, which measures how much more (or less) attention attacker $k$ receives compared to what the defending team typically awards to each attacker. This normalizes attention relative to the team’s own mixture of zonal and man-marking, helping distinguish attackers who consistently draw disproportionate attention across varying defensive schemes.

Let \( A_k^{(p)} \) denote the attention received by attacker \(k\) in corner sequence \(p\), defined as the average number of defenders assigned to mark them per frame over the course of the sequence. Let $\mathcal{P}_k$ be the set of all corner sequences in which attacker $k$ appears, and let $w(p)$ denote the defending team in sequence $p$. For each team $w$, we compute their baseline attention, $\bar A_w$, as the average attention they assign to marked attackers across all sequences they defend:
\[
\bar A_{w} = \frac{1}{|\mathcal{P}_w|} \sum_{p \in \mathcal{P}_w} \left( \frac{1}{|\mathcal{A}_p|} \sum_{\ell \in \mathcal{A}_p} A_\ell^{(p)} \right),
\]
where $\mathcal{P}_w$ is the set of sequences defended by team $w$, and $\mathcal{A}_p$ is the set of attackers in sequence $p$.

The deviation from this baseline for attacker $k$ in sequence $p$ is then defined as
\[
CA_k^{(p)} = A_k^{(p)} - \bar A_{w(p)},
\]
and the final context-aware attention score is the average of this difference across all sequences involving attacker $k$.

\subsubsection*{Attacker Evasiveness}
To assess the evasiveness of attackers making runs into the box during corners, we define an attacker evasion score, $\mathrm{ES}(k)$, which quantifies how much space attacker $k$ gains on their closest marker between the moment he is first man-marked and the moment of first contact with the ball. The score is weighted by how close to goal this separation occurs, attempting to reward the value of creating separation in dangerous areas. The per-sequence scores, $\mathrm{ES}_p(k)$, is defined as:
\[
\mathrm{ES}_p(k) = \phi_{\text{goal}}(t_c) \cdot \Delta d_{p,k}.
\]
Here $\Delta d_{p,k} = d_{\min}(t_c) - d_{\min}(t_0)$ is the change in the minimum Euclidean distance between the attacker and their marker from the first marking frame ($t_0$) to the first contact with the ball ($t_c$). The spatial weighting term, $\phi_{\text{goal}}(t_c)$ is a function of the attacker's distance to goal, ensuring that space gain closer to the goal is valued more highly.

\[
\phi_{\text{goal}}(t_c) = 1 - \frac{\min(18\,\text{yd},\,\mathrm{d^{goal}}_k(t_c))}{18\,\text{yd}}.
\]

\subsubsection*{Interpreting Attacker Profiles}

\begin{figure}[htbp]
    \centering
    \includegraphics[width=\linewidth]{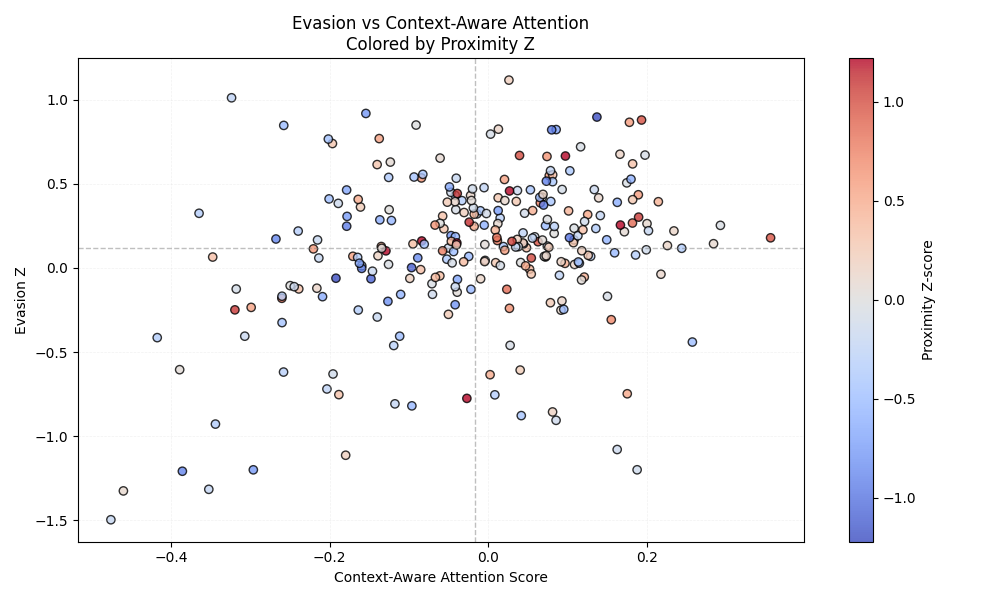}
    \caption{Relationship between attacker evasiveness and context-aware attention during corner kicks in the 2023/24 English Premier League season. Each point represents an individual attacking player who has appeared in at least 20 corner sequences, positioned by their average context-aware attention score (x-axis) and evasion score (y-axis). Colour indicates the player’s first contact proximity, capturing how frequently they operated close to ball delivery location, with warmer tones indicating higher proximity. Players in the top-right quadrant receive above-average defensive attention and create above-average separation from their markers, while those in the bottom-left are less tightly marked and generate limited separation. The remaining quadrants distinguish players who either evade well despite receiving little attention (top-left), or who are marked closely but struggle to create space (bottom-right).}

    \label{fig:Evasion vs Context-Aware Attention}
\end{figure}

We visualize the relationship between context-aware man-marking attention and attacker Separation Score in Figure \ref{fig:Evasion vs Context-Aware Attention} for players who appeared in at least 20 corner sequences. Each point represents an individual player, positioned by their average scores for each metric. The color of each point denotes the player's proximity to the ball at the moment of first contact, with warmer tones indicating those who operate closer to the delivery location. This analysis enables practitioners to quickly identify different types of attacking threats:

\begin{itemize}[leftmargin=2em, rightmargin=2em]
    \item \textbf{Top-right quadrant}: Players with high separation score and high context-aware man-marking attention. These are high-priority threats who are both tightly marked and effective at creating space.
    \item \textbf{Bottom-left quadrant}: Players who are neither tightly marked nor successful at creating space. This suggests they are either not perceived as a primary threat or are not centrally involved in the attacking scheme for this type of set piece.
    \item \textbf{Top-left quadrant}: Players who consistently draw less attention and evade those who do mark them.
    \item\textbf{Bottom-right quadrant}: Players who are often tightly marked and struggle to create space from their marker.
\end{itemize}

\subsubsection*{Effective Number of Initial Man-Marking Assignments}
To assess the diversity of initial man-marking assignments for each defender, we define the effective number of attackers per game, which quantifies how many distinct attackers a defender typically begins marking in a given match. This is calculated by first computing the entropy \cite{ElementsOfInfomationTheory} of a defender’s initial assignments per game, converting it into an effective count of attackers, and then averaging across matches.

For each game $g$, let $C_{j,k}^{(g)}$ denote the number of corner sequences in which defender $j$ begins by marking attacker $k$, and let $T_j^{(g)} = \sum_k C_{j,k}^{(g)}$ be the total number of initial man-marking assignments observed for defender $j$ in that game. The empirical assignment distribution is
\[
p_{j,k}^{(g)} = \frac{C_{j,k}^{(g)}}{T_j^{(g)}},
\]
and the corresponding assignment entropy is given by
\[
H_j^{(g)} = -\sum_k p_{j,k}^{(g)} \log p_{j,k}^{(g)}.
\]
This entropy is converted to an effective number of attackers,
\[
N_j^{\mathrm{eff},(g)} = \exp\left(H_j^{(g)}\right),
\]
which reflects the number of attackers a defender would need to mark uniformly to obtain the same entropy value. The effective number of attackers not only counts how many different opponents a defender ever marks, but also weights them by how evenly that defender’s assignments are distributed. For example, a defender who splits his attention equally between two attackers will have a higher value ($\approx2$) than one who marks one almost exclusively and the other only once ($\approx1.1$), even though both have a unique‐count of 2. The final metric is obtained by averaging across all games in which the defender has at least one observed initial man-marking assignment. Lower values of $N_j^{\mathrm{eff}}$ indicate defenders with consistent initial roles across games, while higher values reflect more varied or flexible marking responsibilities.

\subsubsection*{Switch Rate}
To measure how often defenders change their man-marking responsibilities, we define the Switch Rate. For each defender in a corner sequence, we count the number of times their man-marking assignment changes from one frame to the next. We then divide this count by the total number of frames in the sequence (minus one) to get a per-sequence rate. The final Switch Rate is the average of these per-sequence rates across all corners in which a defender appears, yielding an overall measure that reflects how frequently they transfer between marking different attackers.

\subsubsection*{Interpreting Defender Profiles}

\begin{figure}[htbp]
    \centering
    \includegraphics[width=\linewidth]{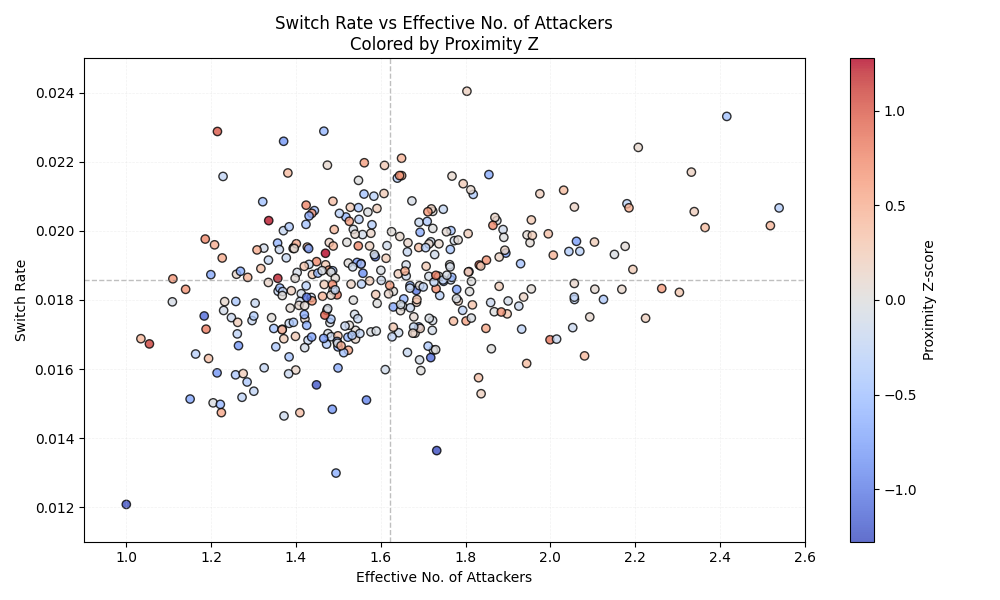}
    \caption{Relationship between defender switch rate and effective number of initial attackers marked, based on data from the 2023/24 English Premier League season. Each point represents a defender who appeared in at least 20 corner sequences, positioned by their average number of distinct attackers marked at the start of corners (x-axis) and their average rate of switching man-marking assignments within sequences (y-axis). Colour denotes the defender's average proximity to the point of first contact, with warmer tones indicating closer positioning. Defenders in the top-right quadrant both frequently switch marking assignments and are assigned to a diverse set of attackers initially, while those in the bottom-left exhibit more stable, specialised marking roles.}

    \label{fig:Switch Rate vs Effective No. of Attackers}
\end{figure}

Figure~\ref{fig:Switch Rate vs Effective No. of Attackers} visualises defender behaviour during corner kicks by plotting the average switch rate against the effective number of inital attackers marked. The effective number of initial attackers captures the variety of initial man-marking assignments a defender receives across games, while the switch rate reflects how often that defender changes assignments within individual sequences. Each point represents a defender who appeared in at least 20 corner routines, and colour denotes their average proximity to the point of first contact, with warmer tones indicating players more frequently located near the delivery zone.

\begin{itemize}[leftmargin=2em, rightmargin=2em]
    \item \textbf{Top-right quadrant}: Defenders who frequently switch assignments and mark a wide variety of attackers. These players may be tasked with flexible or reactive defensive roles.
    \item \textbf{Bottom-left quadrant}: Defenders with a lower effective number of initial assignments and a low switch rate, suggesting a more specialized or rigid role.
    \item \textbf{Top-left quadrant}: Defenders who mark a narrow range of attackers but exhibit high within-sequence switching, which may reflect reactive coverage or difficulty in maintaining individual matchups.
    \item \textbf{Bottom-right quadrant}: Defenders who are assigned to a wide variety of attackers but maintain low within-sequence switching, suggesting their role is to take an initial assignment and maintain it. 
\end{itemize}

\subsection*{Human Analyst Baseline}

To contextualise model performance on corner recipient prediction, we first measured how accurately expert practitioners could anticipate the first-contact recipient from the information available at delivery. Eight football analysts from Nottingham Forest FC (first team and academy staff) were shown a sample of 30 corner routines from our validation set. Each routine was presented as a single image depicting player positions and velocity vectors at the moment of delivery, with teams distinguished by colour and the corner taker highlighted. Participants were instructed to nominate the three players they believed most likely to win first contact.

Across all participants, mean top-3 accuracy was $23.75\% \pm 4.52\%$ (mean $\pm$ standard deviation). This illustrates the inherent difficulty of the task when judged from the delivery snapshot alone, providing a practical reference point for interpreting model performance.

\subsection*{Corner Recipient Prediction}

We compare two graph neural network (GNN) models for predicting which player will next interact with the ball (the first-contact recipient), given a graph representation of the game at the moment of corner delivery. The first model takes canonicalised graphs as inputs, with MLP \cite{rumelhart, Goodfellow2016DeepLearning} pre- and post-processing layers and four intermediate GATv2 \cite{brody2022attentivegraphattentionnetworks, veličković2018graphattentionnetworks} layers. The second model is a reimplementation of TacticAI's receiver prediction component, which employs four D\textsubscript{2}-equivariant GATv2 layers. We evaluate model performance under varying training set sizes corresponding to 50\%, 75\%, and 100\% of available tracking data (roughly 2, 3, and 4 seasons). In each regime, we retain a fixed 20\% test set and split the remainder into training and validation subsets (80/20), subsampling the training set to the target fraction. We report top-3 accuracy in all experiments, following the evaluation protocol used in \cite{RN8}. All models are implemented in PyTorch and trained using the Adam optimizer with early stopping based on validation top-3 accuracy.

At the full dataset size (100\%), the canonicalised model achieved significantly higher average top-3 accuracy (48.07\%) than TacticAI (47.43\%; paired t-test $p=0.0088$). Expressed relative to expert human performance on the same task, these correspond to 202\% (canonicalised) and 200\% (TacticAI) of the human baseline top-3 accuracy. At reduced dataset sizes, differences between models were smaller and not statistically significant (75\%: canonicalised 47.00\%, TacticAI 46.71\%, $p=0.0740$; 50\%: canonicalised 44.81\%, TacticAI 45.22\%, $p=0.2472$). Relative to the human baseline, these correspond to 198\% vs 197\% (75\%) and 189\% vs 190\% (50\%), respectively. Thus, while canonicalisation shows a marginal but statistically significant advantage at full data, performance converges as training data decreases. Notably, despite our models achieving lower headline accuracy than reported in the original TacticAI paper, both models substantially outperform expert human raters on this delivery-time snapshot task.

\subsection*{Evaluation Of Off-ball Defensive Contributions}

\subsubsection*{Defensive Value Added}

To evaluate the efficacy of defensive positioning at the moment of delivery $t$, we compare the reception risk implied by a defender’s observed location to the risk achievable under counterfactual man-marking assignments generated by the CDHMM. Let $a_{tk}$ denote attacker $k$ at time $t$ and $d_{tj}$ denote defender $j$ at time $t$. We use a reception model to estimate the probability that attacker $k$ receives first contact at delivery given defender $j$'s position, written $P_r(a_{tk}\mid d_{tj})$. 

The CDHMM provides a distribution over counterfactual defender positions $x$ induced by assigning defender $j$ to a particular man-marking role $k'$ (i.e., ``mark attacker $k'$''), which we denote $p(x\mid k')$. Under such a role, the expected total reception probability conceded to the feasible attacker set $K_{jt}$ is
\[
\mathbb{E}_{x\mid k'}\!\left[\sum_{k\in K_{jt}} P_r(a_{tk}\mid x)\right],
\]
where $K_{jt}$ is the set of feasible man-marking roles for defender $j$ at time $t$ (equivalently, the attackers considered relevant/threatening for that defender in that frame).  We define the Group Coverage Advantage (GCA) as the gap between (i) the minimum expected group reception probability achievable across feasible counterfactual roles and (ii) the group reception probability implied by the defender’s observed position:
\begin{equation}
\mathrm{GCA}_{tj}
=
\min_{k' \in K_{jt}}
\mathbb{E}_{x\mid k'}\!\left[\sum_{k\in K_{jt}} P_r(a_{tk}\mid x)\right]
-
\sum_{k\in K_{jt}} P_r(a_{tk}\mid d_{tj}).
\end{equation}

If $\mathrm{GCA}_{tj} > 0$ this means that event the best counterfactual man-marking option concedes more than the defender actually conceded, so the defender’s real positioning was better than any of the feasible man-marking alternatives.  By benchmarking against a set of optimal counterfactuals rather than raw reception probabilities, this framework offers several key advantages:
\begin{itemize}[leftmargin=2em, rightmargin=2em]
    \item \textbf{Isolation}: It disentangles the defender's contribution from confounding variables such as passer pressure or the inherent difficulty of the pass, which are held constant across both real and ghost scenarios.
    \item \textbf{Contextual Baselines}: The "optimal" ghost is calculated specifically within the defender’s feasible set $K$, which is itself conditioned on the collective defensive structure, accounting for decision making relative to teammate's positions and roles.
\end{itemize}

\begin{figure}[htbp]
\centering
\includegraphics[width=0.7\textwidth]{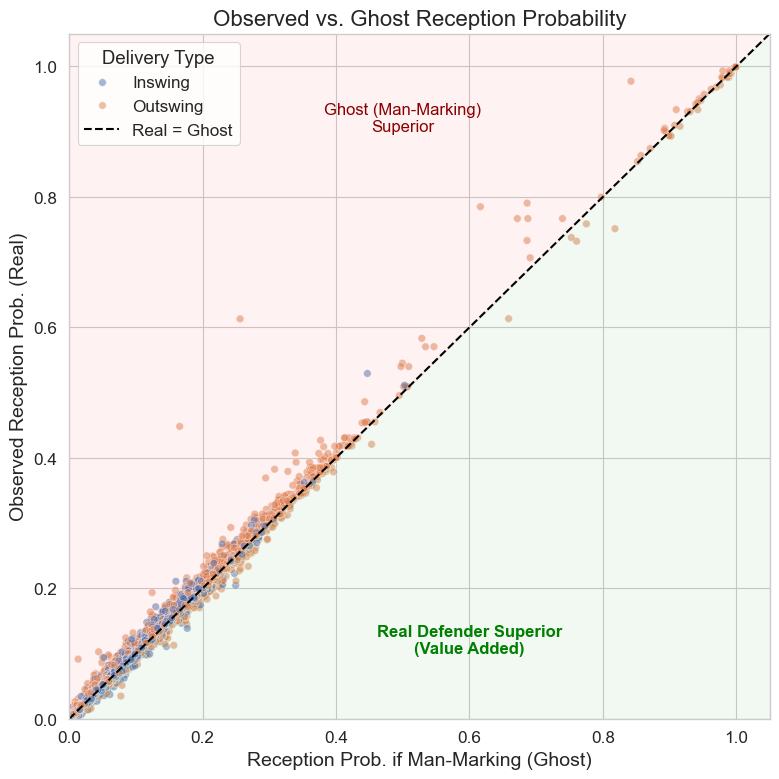}
\caption{Observed vs. Optimal Ghost Reception Probability. The scatter plot compares observed defensive positioning against CDHMM-generated counterfactual "ghosts". For each defender, a set of feasible attackers $K$ is identified based on the CDHMM's expected state occupancy. We calculate the total reception probability for the group $K$ given the observed game state and compare it against the expected outcome if the defender had assumed the optimal man-marking role within that set, determined via Monte Carlo sampling. Points falling below the identity line indicate "Value Added," where the defender's actual positioning suppresses reception probability more effectively than the best average man-marking baseline. Conversely, points above the line signify defensive underperformance relative to the counterfactual baseline.}
\label{fig:ghost_vs_real}
\end{figure}

Figure \ref{fig:ghost_vs_real} shows the majority of actions align with the $y=x$ identity line, suggesting the CDHMM effectively captures standard marking behavior. The "Value Added" region (green) identifies instances where real positioning outperforms the optimal man-marking choice, while the red region serves as a diagnostic tool for defensive underperformance relative to a feasible marking role.

\subsubsection*{Case Studies}

\begin{figure*}[t]
   \centering

   \begin{subfigure}[t]{0.48\textwidth}
      \centering
      \includegraphics[width=\textwidth]{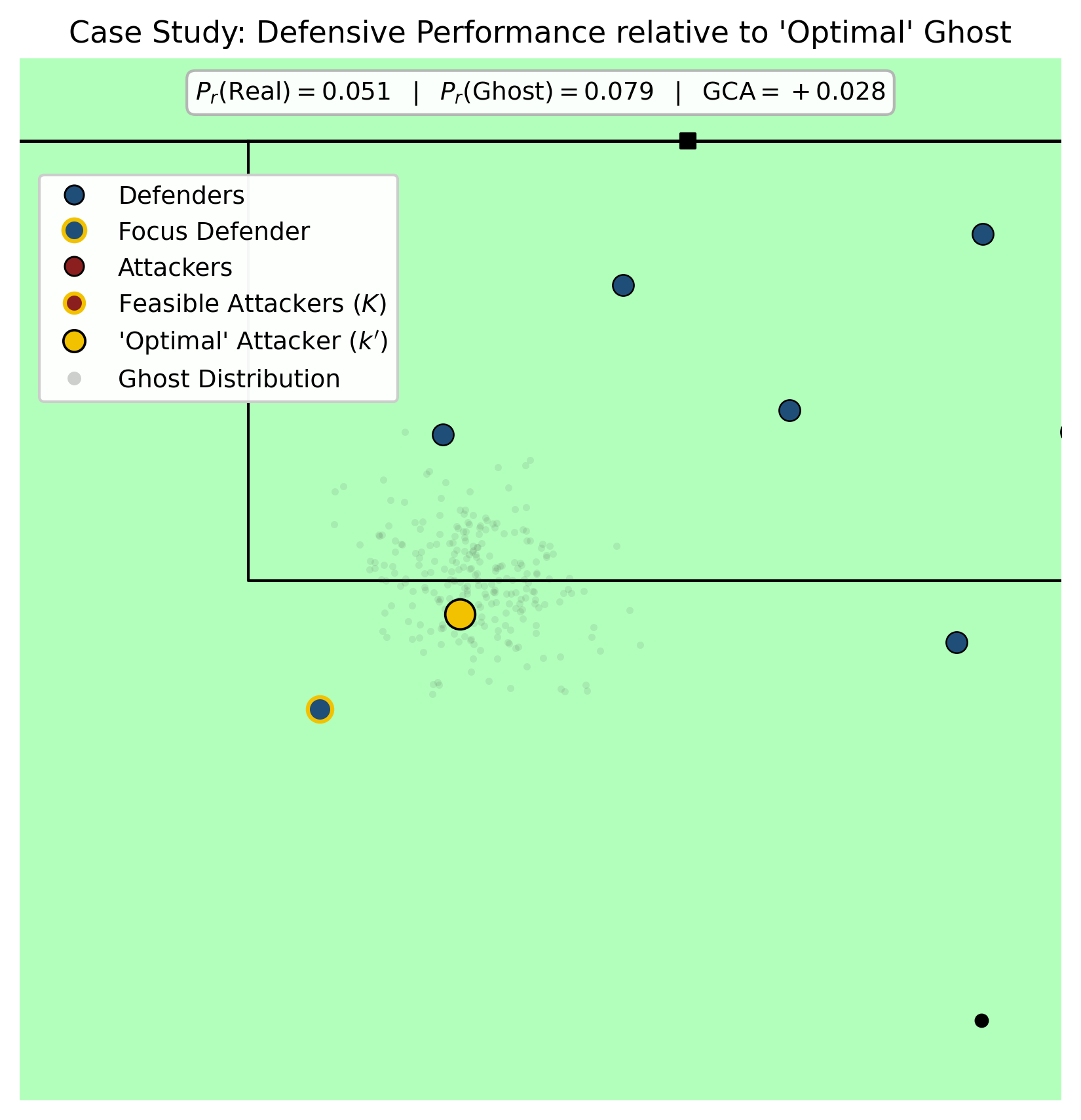}
      \caption{Single-threat scenario ($|K|=1$).\vphantom{with optimal target identification.}}
      \label{fig:case_study_k1}
   \end{subfigure}
   \hfill
   \begin{subfigure}[t]{0.48\textwidth}
      \centering
      \includegraphics[width=\textwidth]{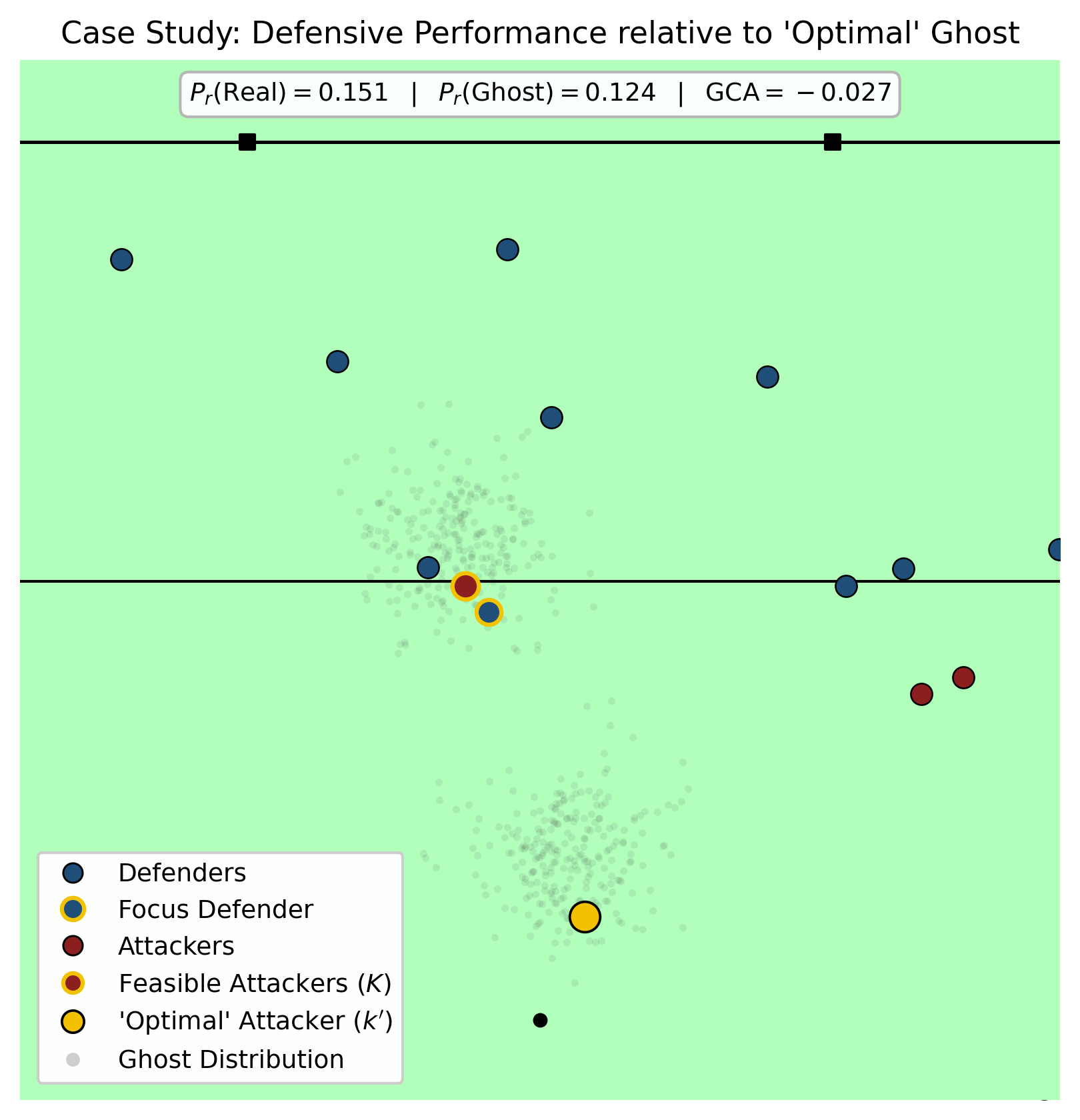}
      \caption{Multi-threat scenario ($|K|=2$) with optimal target identification.}
      \label{fig:case_study_k2}
   \end{subfigure}

   \caption{Spatial Case Studies of GCA. (A) Demonstrates a defender outperforming the ghost baseline by occupying a strategically advantageous zone rather than tight man-marking. (B) Illustrates threat prioritization in an overload; the yellow attacker represents the 'Optimal' target used to calculate the GCA baseline. Real positions are shown in deep blue (defenders) and deep red (attackers), with the counterfactual ghost distribution represented by the grey point cloud.}
   \label{fig:case_studies}
\end{figure*}

The utility of the Group Coverage Advantage (GCA) metric is illustrated through case studies involving varying degrees of attacker complexity ($|K|$). By comparing simple man-marking ($|K|=1$) with complex multi-attacker scenarios ($|K| \geq 2$), we can visualize how defenders prioritize threats.  In Figure \ref{fig:case_studies} A ($|K|=1$), the defender positions themselves relative to a single attacker. The GCA here measures the difference in reception probability for that attacker compared to an "average" marking position, calculated via Monte Carlo sampling (see the point cloud); a negative GCA indicates a defensive lapse where the defender is positioned negatively compared to the optimal man-marking role. Interestingly we see that the model values the defender's positioning despite him not executing a man-marking style of positioning relative to the attacker, showing that given the other defender's positions the defender we are focusing on makes a good decision to position themselves like they do, potentially because they would provide limited value in an area already overloaded by defenders, but by positioning themselves like they do they limit shorter or more front post contact likelihood. Figure \ref{fig:case_studies} B demonstrates a more complex scenario where the defender could choose between $|K|=2$ feasible attackers. Here, GCA is benchmarked against the "Optimal Ghost" man-marking assignment that would minimize the total group reception probability. By highlighting the specific attacker that the GCA is benchmarked against, we demonstrate the model’s ability to identify which threat the defender should have prioritized. In this particular case the model indicates that the defender would reduce the oppositions reception probability most by stepping out from his teammates and towards the optimal attacker who has been left free near the penalty spot. Such examples show that elite defensive value can stem from the ability to "split" the difference between multiple threats or correctly identify the most likely recipient in high-entropy situations.

\subsubsection*{Emergent Defensive Behaviors: Overloads and Delivery Types}

\begin{figure}[htbp]
\centering
\includegraphics[width=0.96\textwidth]{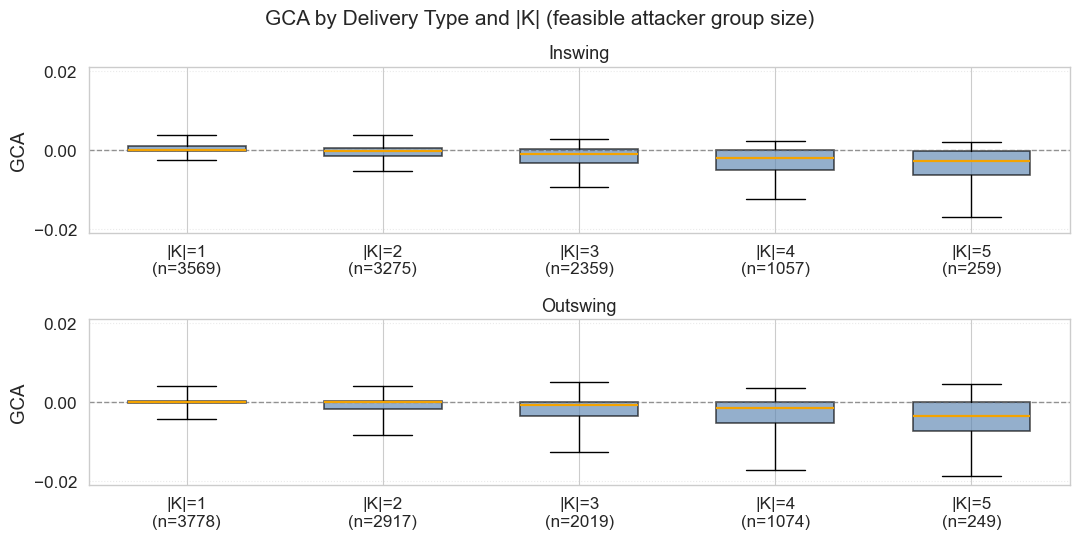}
\caption{Impact of Situational Complexity and Delivery Type on Defensive Performance. Group Coverage Advantage (GCA) is plotted as a function of the number of feasible attackers ($|K|$) for both inswinging (top) and outswinging (bottom) deliveries. As situational complexity increases ($|K|=1$ to $|K|=5$), there is a expansion in the variance and interquartile range of GCA, indicating higher defensive volatility in overload scenarios. While delivery type results in statistically distinct distributional shapes (verified via KS tests), the negligible shift in medians (Cohen’s $d < 0.11$) suggests that the density of the attacking group, rather than ball trajectory, is the primary driver of GCA volatility.}
\label{fig:gca_boxplots}
\end{figure}

To evaluate the impact of delivery trajectory on defensive quality, we analyzed standardized effect sizes (Cohen's $d$) alongside the Kolmogorov-Smirnov (KS) and Mann-Whitney U (MWU) tests. Across all $|K| \leq 4$, Cohen's $d$ remained below 0.11, demonstrating that the practical difference in average Group Coverage Advantage (GCA) between inswinging and outswinging deliveries is negligible. This suggests that ball trajectory does not fundamentally bias a defender's average positioning efficacy.  However, the discrepancy between the MWU and KS test results reveals a shift in defensive consistency. While the MWU test detects a statistically significant but practically minor shift in medians, the KS test yields significantly more extreme $p$-values ($p_{KS} = 1.24 \times 10^{-43}$ vs. $p_{MWU} = 8.96 \times 10^{-17}$ for $|K|=1$). This indicates that the primary effect of delivery type is not on average performance, but on the shape and volatility of the distribution of GCA observed. Specifically, outswinging deliveries appear to create higher-entropy situations, resulting in a wider variance of GCA values—characterized by more frequent elite positioning and more severe defensive lapses—even while average suppression remains stable. This volatility is further amplified by situational complexity. As shown in Figure \ref{fig:gca_boxplots}, the standard deviation of GCA expands as the number of feasible attackers $|K|$ increases. High-complexity 'overload' scenarios thus provide greater opportunities for both high-value defensive interventions and severe failures when seeking to minimise reception probabilities.

\section*{Discussion}

\subsubsection*{CDHMM Outputs and Limitations}

The CDHMM framework presented in this paper enables interpretable, unsupervised analysis of defensive behaviour during corner kicks, uncovering temporally resolved man-marking and zonal assignments. The inferred state sequences, emission distributions, and transition dynamics reflect tactical strategies and also facilitate downstream player-level characterisation. Unlike prior approaches that rely on manually labeled data to identify tactical roles or marking assignments \cite{routineInspection, Bauer}, our unsupervised CDHMM infers these directly from the tracking data. This label-free approach makes the framework scalable and applicable to new teams or scenarios without the need for time-intensive manual annotation. Combined with outcome prediction models such as pass or reception prediction, it facilitates sophisticated analysis of team strategies, and through counterfactual analysis extends the capability of these predictive models to be able to evalulate off-ball defensive contributions. In this section, we reflect on key findings from the model outputs, discuss limitations of the current formulation, and outline directions for future work.

Our analysis of inferred zonal structures revealed systematic differences in spatial organisation between inswinging and outswinging deliveries. For instance, zones estimated for inswinging corners tended to cluster around the near post and six-yard box, while outswinging zones were more dispersed, often extending toward the edge of the penalty area. These patterns are consistent with coaching intuition and tactical adaptations to ball trajectory and delivery type, and suggest that the model captures context-dependent spatial roles without the need for manual annotation.

The distribution of learned transition parameters further supports this interpretation. While self-transition weights for man-marking states were relatively stable across teams and delivery types, parameters governing zonal persistence exhibited greater variability. This highlights zonal marking strategies are more context-sensitive and reflect team-specific defensive schemes. The CDHMM thus captures not just generic defensive tendencies but also nuanced adaptations based on delivery type and team setup.

When we examined player-level behavioural metrics derived from the latent state sequences—such as switch rate and the effective number of initial assignments—we found systematic variation across defenders. Interestingly, defenders appearing in the bottom-right quadrant of Figure~\ref{fig:Switch Rate vs Effective No. of Attackers} were frequently identified as zonal players when discussing model outputs with coaches and analysts. This suggests that zonal defenders tend to begin sequences assigned to a zonal state, and then encounter a variety of attacking opponents across games who enter their zones, but rarely switch assignments within sequences. This likely reflects a stable zone responsibility: attackers enter their area, drawing a transition to man-marking, but once that switch occurs, the assignment typically persists. In contrast, defenders with high switch rates and varied initial assignments appear to take on more reactive or flexible roles. These metrics help distinguish defenders with tightly bounded marking roles from those involved in more complex or adaptive duties.

Despite the interpretability and domain alignment of our CDHMM framework, several limitations remain. First, the model does not currently include an “uninvolved” state to account for defenders who are neither zonal nor actively man-marking. In some sequences where a defender is positioned well away from the active play, or do not consistently adopt a similar team level zonal position, this can lead to forced assignments that do not match intuitive roles. Second, the duration of time spent in each hidden state is implicitly governed by geometric distributions due to the standard HMM formulation. This may underestimate longer-term stability in assignments. Third, the CDHMM does not utilise observed player velocities in its emission distribution formulation, extending the model to do so could potentially improve the models ability to detect zonal and man-marking behaviours. Regarding our emission model, the man-marking formulation is tailored for set-piece behaviour and may not generalise well to open play, where player movements are more variable and player's reference frames for positioning are less stable. Similarly, the zonal emission distributions would require revision, as zones in open play are more dynamic and often defined by relative positioning to teammates rather than static locations adopted at the start of a sequence. Assuming zonal distributions are Gaussian imposes symmetry around the zone mean, which may not hold in practice. For example, defenders may consistently bias their positioning toward the goal or corner spot, resulting in skewed distributions that the model cannot capture.

\subsubsection*{Off-ball Contribution Analysis}

To our knowledge, this work represents the first attempt in football to generate counterfactual representations of play that are explicitly conditioned on interpretable defensive behaviours. By modelling who is marking whom—and how these assignments evolve over time—we enable tactical diagnosis and support simulations that "ghost" alternative defensive structures. These counterfactuals can be passed through outcome prediction models, allowing us to investigate the effect of individual defenders on predicted outcomes such as receptions or expected threat. This enables new methods for quantifying off-ball contributions, either by comparing a defender’s observed behaviour to plausible alternatives or by generating defender-specific weights for credit assignment. It is important to note that our evaluation does not compare each defender to an optimal positioning policy that globally minimises opponent threat. Instead, we assess their contribution relative to a counterfactual in which they strictly mark each attacker within a plausible spatial neighbourhood. More complex comparisons could be facilitated in future work using grid searches over defender positions and velocities, or by employing reinforcement learning agents trained to suppress opponent threat—an approach analogous to recent work comparing passing decisions to those made by learned agents \cite{RN1}.

\subsubsection*{HMM Future Work}

Future work will expand this framework into open play analysis. While set pieces offer structure and repeatability, many of the underlying questions—who is responsible for preventing progress, how defenders adapt to off-ball movement—are equally relevant in dynamic play. Overall, the CDHMM provides a novel, interpretable approach to understanding team and individual defensive behaviours from tracking data, and lays the foundation for further integration of unsupervised behavioural models with predictive analytics in football.

\subsubsection*{Reception Prediction Models}

Our first recipient prediction models achieve top-3 accuracies between 44–48\%, this compares favourably to a baseline accuracy of 23.75\% obtained from expert analysts using static tracking data visualisations, highlighting both the difficulty of the task and the value of predictive modelling in supporting decision-making under uncertainty.

The marginally superior performance of the canonicalised model over the D\textsubscript{2}-equivariant TacticAI model, particularly at full dataset size, suggests that the assumptions underlying equivariance to D\textsubscript{2} symmetries may not hold perfectly in the football corner-kick prediction scenarios. Football set pieces, such as corner kicks, inherently exhibit asymmetries due to teams frequently assign predominantly left or right-footed players to take corners on particular sides of the pitch. Because there is a higher fraction of players who are predominantly right-footed compared to left-footed this may break the assumption of perfect mirror symmetry that D\textsubscript{2}-equivariant architectures exploit. Consequently, enforcing equivariance may unintentionally restrict the representational capacity or flexibility of the model, potentially limiting its predictive accuracy.

The improved performance of the canonicalised model at full dataset size might be attributable to its increased modeling capacity, provided by additional MLP preprocessing layers. These layers could facilitate learning more expressive  representations beyond symmetry constraints. However, this benefit diminishes with reduced dataset sizes, as indicated by the lack of statistical significance at 75\% and 50\% data conditions, suggesting that the additional capacity may not be fully utilized or could even lead to mild overfitting when fewer training samples are available. In scenarios with limited data, the TacticAI model’s equivariant constraints could potentially offer regularisation benefits, preventing overfitting. However, our results do not clearly support this hypothesis, as we observed no significant performance advantage for the equivariant model at smaller dataset sizes.

\subsubsection*{Limitations and Dataset Differences}

While our reimplementation of TacticAI’s receiver prediction model achieves top-3 accuracy of approximately 47\%, the original paper reports values closer to 75\%. Our implementation closely follows the released code and uses the same architecture and D\textsubscript{2}-equivariant GNN layer formulation. Although both models operate on Premier League tracking data the datasets are not strictly identical. In particular, the process by which corner sequences are extracted and aligned to ball delivery events may differ. Clubs often maintain proprietary event alignment pipelines, and discrepancies in how “moment of delivery” is defined could result in subtle but important task differences. For example, if our sequences consistently begin earlier relative to the actual ball contact, our models may face a more difficult prediction task. To investigate this, we visually inspected hundreds of our aligned sequences alongside game video and found no evidence of systematic misalignment. Our event alignment appears well-behaved, though differences in annotation protocols remain a plausible explanation for the performance gap. Consequently, while our TacticAI reimplementation is faithful at the architectural level, our results should be interpreted as a comparative benchmark rather than a direct replication of the original.

\section*{Methods}

\subsection*{CDHMM Definitions}

We utilize an CDHMM, an extension the standard HMM \cite{Bishop}, where transition probabilities vary with time-varying covariates. 
For each team and delivery type, we model a defender $j \in \{1,..,J\}$ with $N=K+1$ latent behavioral states, where $J$ is the number of outfield defenders and $K$ is the number of outfield attackers. The hidden states $s_t \in  \mathcal S=\{q_1,\dots,q_K,q_N\}$ represent man-marking an attacker $k$ and defending an assigned zone, respectively. We set $J=K=10$. Each defender is treated independently, except during initial zone assignments. 

The model defines $z \in \{1,...,10\}$ for each team and delivery type. If defender $j$ is in the zonal state, $s_t = q_N$, their position, $D_{t,j}$ at time $t$ is governed by an emission distribution specific to their assigned zone: 
\[
  D_{t,j}\sim\mathcal N(\mu_z,\Sigma_z),
\]
where $\mu_z$ and $\Sigma_z$ are the mean and covariance of zone $z$. At the start of a sequence ($t=0$), defenders are uniquely assigned to zones by solving a linear sum assignment problem using the Kuhn-Munkres algorithm \cite{Kuhn, Munkres} to minimize the sum of Euclidean distances between each defender's position and an assigned zone's mean:
\[
  \min_{X\in\{0,1\}^{K\times K}}\sum_{j=1}^K\sum_{z=1}^K \|D_{0,j}-\mu_z\|_1\,X_{j,z}
  \quad
  \text{s.t.}\quad \sum_z X_{j,z}=1,\;\sum_j X_{j,z}=1.
\]

To model man-marking behavior, we divide the pitch into fixed \(3\,\mathrm{m} \times 3\,\mathrm{m}\) spatial bins, indexed by \(l\). Although Premier League pitch sizes vary slightly, penalty box dimensions are standardized; we therefore define bins with fixed physical dimensions to ensure consistency in the regions surrounding the penalty area. For each bin \(l\), we define a unique parameter vector \(\Gamma^{(l)} = [\gamma_o^{(l)},\, \gamma_g^{(l)}]\) and variance \(\sigma^{2(l)}\), subject to the constraint \(\gamma_o^{(l)} + \gamma_g^{(l)} = 1\). This parameterization ensures the defender's expected position lies on the line connecting their assigned attacker and the center of the goal. Given that defender \(j\) is in state \(s_t = q_k\), i.e., marking attacker \(k\), and attacker \(k\)'s position \(O_{t,k}\) lies within bin \(l\), the defender’s position \(D_{t,j}\) is modeled as
\[
D_{t,j} \sim \mathcal{N}(\mu_{t,k}^{(l)}, \sigma^{2(l)}),
\quad\text{where}\quad
\mu_{t,k}^{(l)} = \gamma_o^{(l)} O_{t,k} + \gamma_g^{(l)} G,
\]
and \(G\) denotes the center of the defending team’s goal line. The bin-specific weights, $\gamma_o^{(l)}$ and $\gamma_g^{(l)}$, control how tightly a defender marks their opponent, reflecting the positional strategy in that area of the pitch.

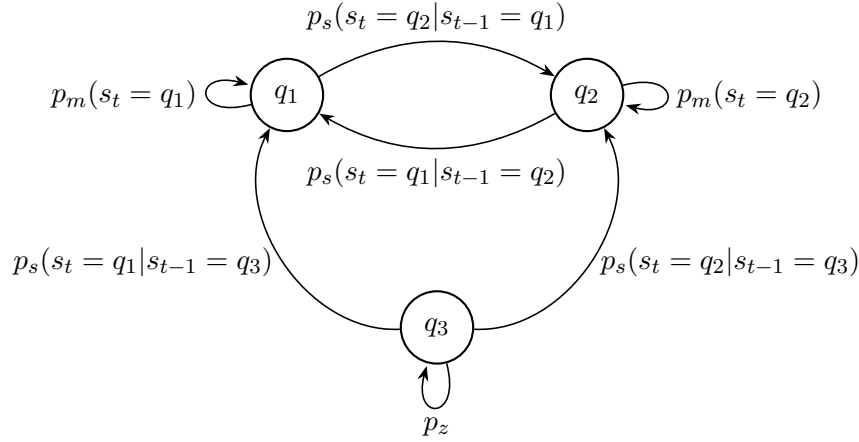
\begin{figure}[tb] 
    \centering
    \begin{tikzpicture}[->, >=Stealth, auto, semithick, node distance=2cm]
    \tikzset{every state/.style={fill=white, draw=black, thick, text=black, scale=1}}

    \node[state] (q1) {$q_1$};
    \node[state] (q2) [right=3cm of q1] {$q_2$};
    \node[state] (q3) [below=2.598cm of $(q1)!0.5!(q2)$] {$q_3$}; 

    \path
    (q1) edge[loop left] node{$p_m(s_t=q_1)$} (q1)
         edge[bend left, above] node{$p_s(s_t=q_2|s_{t-1}=q_1)$} (q2)
    (q2) edge[loop right] node{$p_m(s_t = q_2)$} (q2)
         edge[bend left, below] node{$p_s(s_t=q_1|s_{t-1}=q_2)$} (q1)
    (q3) edge[bend left=60] node[left]{$p_s(s_t=q_1|s_{t-1}=q_3)$} (q1)
         edge[bend right=60] node[right]{$p_s(s_t=q_2|s_{t-1}=q_3)$} (q2)
         edge[loop below] node{$p_z$} (q3);
    \end{tikzpicture}
    \caption{A simplified Markov chain diagram with three states illustrating transition probabilities. This represents the Markov chain for an individual defender in the case where there are only two outfield players ($K=2$) that the defender can mark.  States $q_1$ and $q_2$ are man-marking states and $q_3$ is a zonal state.}
    \label{fig:markovchain}
\end{figure}

Our transition model defines four types of transitions a defender can make between consecutive frames: a man-marking self-transition, a man-marking switch, a zonal self-transition, or a zonal-to-man transition. Based on tactical guidance from coaches and analysts, we prohibit transitions from man-marking to zonal defending.

We parameterize these transitions using logistic and softmax functions over covariate vectors. Separate covariate vectors and weight parameters are defined for man-marking self-transitions ($\beta_m$) and zonal self-transitions ($\beta_z$). Transitions involving a switch between states—either between two attackers or from zonal marking to man-marking—share a single covariate vector and a weight vector ($\beta_s$).  This shared parameterization reflects the conceptual similarity between these switching behaviors and is also statistically motivated, as such transitions occur relatively infrequently and pooling them improves estimation stability.

The probabilities for each transition are defined as follows:
\begin{itemize}
    \item \textbf{Continuing to man-mark $k$}: This is modeled with a sigmoid function, where $X_{t-1,j}^{(m,k)}$ is the covariate vector. 
\[
P(s_{t,j} = q_k \mid s_{t-1,j} = q_k) = \sigma\left(\beta_m^\top X_{t-1,j}^{(m,k)}\right) \equiv p_{m,j,k},
\]
    \item \textbf{Switching from attacker $k'$ to attacker $k$ ($k\neq k'$)}: This probability is proportional to the likelihood of not continuing to man-mark $k$, with probability mass being distributed amongst other attackers via a softmax function.
\[
P(s_{t,j} = q_k \mid s_{t-1,j} = q_{k'},\;k \neq k') = \left(1 - p_{m,j,k'}\right) \cdot \frac{\exp\left(\beta_s^\top X_{t-1,j}^{(s,k)}\right)}{\sum_{\ell \neq k'} \exp\left(\beta_s^\top X_{t-1,j}^{(s,\ell)}\right)} \equiv p_{s,j}(k', k).
\]
    \item \textbf{Transitioning from man-marking to zonal defending}: This transition is disallowed.
\[
P(s_{t,j} = q_N \mid s_{t-1,j} = q_k) = 0 \quad \text{for all } k \in \{1, ..., K\}.
\]
    \item \textbf{Remaining in the zonal state}: This is also modeled with a sigmoid function, where $X_{t-1,j}^{(z)}$ is the covariate vector.
\[
P(s_{t,j} = q_N \mid s_{t-1,j} = q_N) = \sigma\left(\beta_z^\top X_{t-1,j}^{(z)}\right) \equiv p_{z,j}.
\]
    \item \textbf{Transitioning from zonal to man-marking attacker} $k$: This probability is conditioned on not remaining in the zonal state.
\[
P(s_{t,j} =q_ k \mid s_{t-1,j} = q_N) = \left(1 - p_{z,j}\right) \cdot \frac{\exp\left(\beta_s^\top X_{t-1,j}^{(s,k)}\right)}{\sum_{\ell = 1}^{K} \exp\left(\beta_s^\top X_{t-1,j}^{(s,\ell)}\right)} \equiv p_{s,j}(N, k).
\]
\end{itemize}






\subsection*{Covariate Design and Feature Scaling}
All covariate vectors are standardized (zero mean, unit variance) across the dataset to ensure consistent scaling. These vectors are designed to capture a range of dynamic and physical factors that influence a defender's tactical decisions. To quantify whether a defender-attacker pair is approaching or moving apart, we use a convergence metric, included in the covariates for man-marking and switching states:

\begin{equation}
\text{Convergence}_{t,j,k} = \frac{(\mathbf{v}_{t,k} - \mathbf{v}_{t,j})^\top (\mathbf{p}_{t,k} - \mathbf{p}_{t,j})}{\|\mathbf{p}_{t,k} - \mathbf{p}_{t,j}\| + \varepsilon},
\end{equation}

\noindent where $\mathbf{p}_{t,j}$ and $\mathbf{v}_{t,j}$ are the position and velocity of defender $j$ at time $t$, and $\varepsilon$ is a small constant for numerical stability. 

To capture lateral movement, we define the tangential relative velocity as the magnitude of the component of relative velocity that is orthogonal to the defender-attacker line. This metric helps distinguish lock-step marking from situations where an attacker is "slipping past" the defender:

\begin{equation}
v^{\perp}_{t,j,k} = \left\| (\mathbf{v}_{t,k} - \mathbf{v}_{t,j}) - \left[ (\mathbf{v}_{t,k} - \mathbf{v}_{t,j}) \cdot \hat{\mathbf{r}}_{t,j,k} \right] \hat{\mathbf{r}}_{t,j,k} \right\|, \quad
\hat{\mathbf{r}}_{t,j,k} = \frac{\mathbf{p}_{t,k} - \mathbf{p}_{t,j}}{\|\mathbf{p}_{t,k} - \mathbf{p}_{t,j}\| + \varepsilon}.
\end{equation}

We also include the heading alignment, defined as the cosine of the angle between defender and attacker velocity vectors:

\begin{equation}
\cos\theta_{t,j,k} = \frac{\mathbf{v}_{t,j}^\top \mathbf{v}_{t,k}}{\|\mathbf{v}_{t,j}\| \, \|\mathbf{v}_{t,k}\| + \varepsilon}.
\end{equation}

The covariate vector $X_{t,j}^{(z)}$ only contains the convergence metric because zones are stationary.  Additionally, we use player heights and weights to capture physical mismatches. For zonal defending, we use Mahalanobis distance to account for spatial uncertainty, as the size and shape of team zones vary. All covariates are summarized in Table~\ref{tab:feature_vectors}.

\begin{table}[h]
\centering
\caption{Feature composition of covariate vectors used in the transition models.}
\begin{tabular}{l p{7cm} p{5cm}}
\toprule
\textbf{Covariate Vector} & \textbf{Features} & \textbf{Related Transitions} \\
\midrule
$X_{t,j}^{(m,k)}$ & 
1 (bias term), distance from defender to attacker, inverse distance to attacker, tangential relative velocity, heading alignment, convergence metric, attacker height, attacker weight &
Defender $j$ continues to man-mark attacker $k$ \\
$X_{t,j}^{(z)}$ & 
1 (bias term), Mahalanobis distance to assigned zone, inverse Mahalanobis distance, defender velocity toward zone, defender height, defender weight &
Defender $j$ continues to defend zonally \\
$X_{t,j}^{(s,k)}$ & 
1 (bias term), distance to attacker, inverse distance, tangential relative velocity, heading alignment, convergence metric, attacker height, attacker weight &
Defender $j$ switches to man-mark attacker $k$ not previously assigned at $t-1$ \\
\bottomrule
\end{tabular}
\label{tab:feature_vectors}
\end{table}

\subsection*{CDHMM Training}

Model parameters, $\theta=\{\Gamma, \sigma^2, \mu_z, \sigma_z^2, \beta_m,\beta_z,\beta_s\}$, are estimated using an expectation maximisation approach as outlined in algorithm 1. All results are based on training a batch of 10 models per team and delivery type, each for 15 fixed iterations. For each batch, the model with the highest total observation likelihood was selected. We denote the set of $P$ observation sequences as $\mathbf{O} = [\mathbf{O^{(1)}}, ... , \mathbf{O^{(P)}}]$ where $\mathbf{O}^{(p)} = [\mathbf{O}_1^{(p)}, ... , \mathbf{O}_{T_p}^{(p)}]$. Here, $\mathbf{O}^{(p)}$ represents the sequence of tracking data for an individual corner kick sequence $p$, which consists of $T_p$ frames. In the EM algorithm we alternate between computing the posterior expected transition counts
$\xi_t(i, j)=P\bigl(s_t=i,s_{t+1}=j\mid \mathbf{O},\theta)$ and expected state occupancy counts $\gamma_t(i) = P(s_t=i\mid \mathbf{O},\theta)$ using the Forward-Backward algorithm \cite{Bishop}
(E‐step), and then maximizing the expected complete‐data log‐likelihood
(M‐step).

First, \( K \) zone locations are randomly initialized by sampling from a uniform distribution along the six-yard line in the y-direction and extending five meters on either side of the six-yard line in the x-direction. Each zone is initialised with a covariance $\sigma_z^2 = 2I$, where \(I\) represents the identity matrix. The parameters \( \Gamma^{(l)} \) are initialized as \([0.8, 0.2]\) for each bin \( l \). All beta weights are randomly initialised by drawing random values from a normal distribution with zero mean and a standard deviation of 0.1. All initial state probabilities are uniformly set to $1/K$, ensuring that each state has an equal probability at the start.

Treating each defender independently, expected state occupancy and transition counts are calculated using the Forward-Backward algorithm. The observation likelihood of an individual defender in the sequence $p$, $L_{pj} = P(\mathbf{O}^{(p)}|\lambda, j)$, is calculated using the forward-backward algorithm. The complete observation log likelihood is therefore $$\sum_{p,j,n}\log{\alpha_{j,T_p}(n)},$$ where $\alpha_{j,T_p}(n)$ is the forward probability at time $T_p$ for defender $j$ whilst they occupy state $n$ in corner sequence $p$. For each team $K$ zonal locations are estimated, each zone position $\mu_z$ is estimated via $$\hat{\mu_z} = \frac{\sum_{p=0}^P \frac{1}{L_{pj}} D_{0j} \gamma_{0jN}}{\sum_{p=0}^P \frac{1}{L_{pj}} \gamma_{0jN}},$$ where $D_{0j}$ is the position of defender $j$ who is assigned to the zone at time $t=0$. $\gamma_{0jN}$ is the expected state occupancy of defender $j$ occupying their assigned zonal state at time $t=0$ in corner sequence $p$. The zonal distribution has covariance which is estimated using $$\hat{\sigma_z^2} = \frac{\sum_{p=0}^{P}\frac{1}{L_{pj}} \sum_{t=0}^{T_p} \gamma_{tjN} (D_{tj} - \hat{\mu_z})(D_{tj} - \hat{\mu_z})^\top}{\sum_{p=0}^{P}\frac{1}{L_{pj}} \sum_{t=0}^{T_p} \gamma_{tjN}}.$$

\begin{algorithm}[t]
    \caption{Algorithm for HMM Training}
    \label{alg:hmm_training}
    \begin{algorithmic}[1] 
        \STATE \textbf{Initialisation:}
        \STATE Initialise zonal distributions
        \STATE Initialise all bin's $\Gamma^{(l)}$ 
        \STATE Initialise transition weights $\beta_m, \beta_z, \beta_s$
        \STATE Initialise initial state distribution
        \WHILE{not converged}
            \STATE \textbf{Expectation Step (E-step):}
            \FOR{each corner sequence $O_p$ in $O = [O_1, \dots, O_P]$}
                \STATE Assign defending players to zones using the Hungarian algorithm.
                \FOR{each outfield defending player}
                    \STATE Compute forward and backward probabilities using current HMM parameters
                    \STATE Compute sequence observation likelihood $L_{pj}$
                    \STATE Compute expected state occupancy and expected state transition counts
                    \STATE Compute hidden state sequence using the Viterbi algorithm
                \ENDFOR
            \ENDFOR
            \STATE \textbf{Maximization Step (M-step):}
            \STATE Estimate new zone positions and covariances
            \STATE Update transition weights $\beta_m, \beta_z, \beta_s$
            \STATE Update initial state distribution
            \FOR{each man-marking bin $l$}
                \STATE Estimate $\Gamma^{(l)}$ and $\sigma^{2(l)}$
            \ENDFOR
            \STATE Check number of iterations or for convergence criteria
        \ENDWHILE
    \end{algorithmic}
\end{algorithm}

\noindent To estimate $\Gamma^{(l)}$ and $\sigma^{2(l)}$ for each bin $l$, we define the design matrix $\mathbf{Z}^{(l)}$ as follows:
\[
\mathbf{Z}^{(l)} = \bigoplus_{\substack{k \in \text{Attackers} \\ \text{dist}(k, l) \leq N_H}} [O_{tk}, G],
\]
where $\bigoplus$ denotes the concatenation operation, $O_{tk}$ represents the position of attacker $k$ that the defender is marking at time $t$, and $G$ is the goal position. The matrix includes entries for attacker $k$ who is positioned in bin $l$ or in any neighbouring bin within $N_H$ hops of bin $l$. In this analysis, we set the parameter $N_H$ equal to one so only immediate neighbours are included. Any player whose position falls outside the defined bins is included in the nearest possible bin. With our modified design matrix, in the M-step we update $\Gamma^{(l)}$ and $\sigma^{2(l)}$ as outlined by Franks et al.,

\hspace{0.3cm}

$$(\hat{\Gamma}^{(l)}, \hat{\sigma}^{2(l)}) \leftarrow \underset{\Gamma^{(l)}, \sigma^{2(l)}}{\arg\max} \sum_{t,j,k} \frac{\gamma_{tjk}}{\sigma^{2(l)}} (D_{tj} - \Gamma^{(l)} Z^{(l)}_{tk})^2,$$

where $\gamma_{tjk}$ is the expected state occupancy for defender $j$ occupying man-marking hidden state $k$ at time $t$.  Initial state probabilities for each defender are updated using the expected state occupancy values $\gamma_{0jn}$.

In a HMM, the complete-data log-likelihood $Q$ (the log of the joint probability of the hidden state sequence and the observed data) decomposes into two independent terms, one involving the emission parameters and one involving the transition parameters. Consequently, in the M-step of the EM algorithm each parameter set can be maximized separately, without affecting the other \cite{Bishop}, such that

\[
  \beta^{(new)}
  \;=\;
  \arg\max_{\beta}
  \;Q\bigl(\beta\mid\beta^{(old)}\bigr).
\]

Because we assume defenders act independently, the full expected complete‐data log‐likelihood factorizes into a sum over defenders and time steps. Including $l^2$ normalisation, an individual defender's Q function corresponding to transitions is

\begin{equation}\label{eq:Q_complete}
\begin{aligned}
Q(\beta_m,\beta_z,\beta_s)
&=\sum_{t=1}^{T-1}\Biggl\{
   \sum_{i=1}^{K}
     \xi_t(i, i)\,\log p_m
   +\sum_{i=1}^{K}\sum_{\substack{l=1\\l\neq i}}^{K}
     \xi_t(i,l)\,
     \log\bigl[(1 - p_m)\,p_s(i, l)\bigr]\\[-0.3em]
&\quad\;+\;\xi_t(N, N)\,\log p_z^{}
   +\sum_{i=1}^{K}
     \xi_t(N, i)\,
     \log\bigl[(1 - p_z)\,p_s(N, i)\bigr]
\Biggr\}\\
&\qquad
   -\;\tfrac12\Bigl(
       \lambda_m\|\beta_m\|^2
     + \lambda_z\|\beta_z\|^2
     + \lambda_s\|\beta_s\|^2
   \Bigr).
\end{aligned}
\end{equation}

By using the identity $\sum_{l=1}^{K}\xi_t(i, l)=\gamma_{ti}$ which follows from marginalizing over next states, we immediately get $
  \sum_{l\neq i}^{K}\xi_t(i,l)
  =\gamma_{ti}-\xi_t(i,i)$.
It then follows that the gradients of the \(Q\)–function are
\begin{equation}\label{eq:grad_beta_m_Q}
\begin{aligned}
\nabla_{\beta_m}Q 
= \sum_{t=1}^{T-1}\sum_{i=1}^{K}
   \bigl[\xi_t(i,i)-\gamma_{ti}\,p_m\bigr]\;
   X_t^{(m,i)}
\;-\;\lambda_m\,\beta_m,
\end{aligned}
\end{equation}

\begin{equation}\label{eq:grad_beta_z_Q}
\begin{aligned}
\nabla_{\beta_z}Q
= \sum_{t=1}^{T-1}
   \bigl[\xi_t(N,N)-\gamma_{tN}\,p_z\bigr]\;
   X_t^{(z)}
\;-\;\lambda_z\,\beta_z,
\end{aligned}
\end{equation}

\begin{equation}\label{eq:grad_beta_s_Q}
\begin{aligned}
\nabla_{\beta_s}Q
= \sum_{t=1}^{T-1}\Biggl\{
   \sum_{i=1}^{K}\sum_{\substack{l=1\\l\neq i}}^{K}
     \xi_t(i,l)\Bigl[X_t^{(s,l)}
     -\sum_{\substack{k=1\\k\neq i}}^{K}p_s(i, k)\,X_t^{(s,k)}\Bigr]
   \;
\\[-0.3em]+\;
   \sum_{l=1}^{K}
     \xi_t(N,l)\Bigl[X_t^{(s,l)}
     -\sum_{k=1}^{K}p_s(N, k)\,X_t^{(s,k)}\Bigr]
\Biggr\}
\;-\;\lambda_s\,\beta_s.
\end{aligned}
\end{equation}

For each vector of transition weights $\beta$, we compute its gradient and then numerically minimize the negative of our Q-function with respect to $\beta$ using the limited-memory BFGS algorithm with bound constraints (L-BFGS-B), capping the procedure at 100 iterations and utilising normalisation penalties of $\lambda_m=100, \lambda_z = 100, \lambda_s=1000$.

\subsection*{Graph Neural Network Models}

To predict which player will make first contact with the ball, we implemented two supervised graph neural network (GNN) architectures. Both models operate on a static graph representation of the match state at the moment of delivery. In this graph, all on-pitch players are represented as nodes in a fully connected, undirected graph. Each node encodes individual player information, and edges represent the pairwise relationship of being on the same or opposing team.

As detailed in Table \ref{tab:gnn_features}, each node's input features include seven attributes describing the player's position, velocity, physical characteristics, and role as the corner taker. Edges are encoded with a single binary feature to indicate whether the connected players are teammates or opponents. We did not use any global graph-level features.

\begin{table}[htbp]
  \centering
  \caption{Graph input features for first-contact prediction.}
  \label{tab:gnn_features}
  \begin{tabular}{lll}
    \toprule
    \textbf{Level} & \textbf{Feature} & \textbf{Description} \\
    \midrule
    Node           & $x, y$          & Player position (m) on standardised pitch \\
                   & $v_x, v_y$      & Instantaneous velocity (m/s) \\
                   & Height          & Player height (m) \\
                   & Weight          & Player weight (kg) \\
                   & Is corner taker bool       & Boolean indicator \\
    \midrule
    Edge           & teammate bool       & 1 if same team, else 0 \\
    \bottomrule
  \end{tabular}
  \label{tab:Graph features}
\end{table}

\subsubsection*{Equivariant GNN Architecture: TacticAI-Inspired Model}

The first architecture follows the design introduced by Wang et al. in the TacticAI system for corner kick analysis \cite{RN8}. It combines a GATv2-based message-passing architecture with equivariant group convolutions to utilise the reflective symmetries as an inductive bias. The graph is first duplicated across four views corresponding to the D\textsubscript{2} symmetry group: the original layout, horizontal reflection, vertical reflection, and both reflections. Each view is processed with group-equivariant layers and their embeddings are averaged to obtain a reflection-invariant representation of shape \(22 \times 4\). A final linear layer maps each node embedding to a logit, and the model is trained using a weighted cross-entropy loss to address the class imbalance between the single contact-making player and the 21 non-contact players. We apply a weight of \(\sqrt{21}\) to the negative class. We performed a hyperparameter search across a range of values, including learning rate \(\{1\times10^{-4},\,3\times10^{-4},\,1\times10^{-3}\}\), weight decay \(\{1\times10^{-4},\,1\times10^{-3},\,2\times10^{-3}\}\), and dropout: \(\{0.1,\,0.2,\,0.4\}\).

\subsubsection*{Canonicalised MLP + GATv2 Model}

We also implement a non-equivariant GNN model using canonicalised corner frames, in which all corners are rotated and reflected to appear as if taken from the same pitch quadrant. Each node's raw features are first passed through a two-layer multilayer perceptron (MLP) to create a hidden representation. The resulting embeddings are then processed by four standard (non-equivariant) GATv2 layers which use the same number of attention heads (8) and the same per-head output dimension (4) as in the TacticAI-based model. A second two-layer MLP maps the final embeddings to logits for 22-way classification and same weighted cross-entropy loss and hyperparameter grid are used.



\subsection*{Evaluating Off-Ball Defensive Contributions through Counterfactual Analysis and Defensive Credit Attribution}

\subsubsection*{Off-Ball Pass Restriction}
As a first demonstration of our framework's utility, we define the Off-Ball Pass Restriction metric (OBPR), which uses the expected state occupancy ($\gamma_{tjk}$) from the CDHMM to allocate credit to individual defenders. This metric rewards defenders for restricting the pass reception probability of attackers they are marking.
\begin{equation}
\text{OBPR}_j = \sum_{t,k} \gamma_{tjk} \cdot \left(1 - P_r(a_{tk})\right),
\end{equation}
where $P_r(\cdot)$ is pass reception model, returning the player-level probability that a given player will be the next to interact with the ball and $a_{tk}$ is the observed position of attacker $k$ at time $t$.
This metric measures how effectively a defender suppresses passing options across all possible marking assignments, with contributions weighted according to their expected responsibility for each attacker. Similar metrics could be defined by substituting $P_r(\cdot)$ with $P_{\text{threat}}(\cdot)$, an expected threat model, returning the player-level probability that the sequence of play results in a goal for the team conditioned on a player receiving a successful pass. A more complete metric would use both such models jointly to appraise how defenders restrict passes to highly threatening attackers.

Despite awarding credit to individual defenders, this metric is limited in several ways. First, it does not address defensive or offensive overloads—situations where multiple defenders contribute to suppressing the same attacker, or where a defender’s positioning fails to restrict the most threatening of multiple nearby attackers. Secondly, it does not incorporate any situation-specific baseline that accounts for confounding effects, such as on-ball pressure affecting the passer (not the recipient) and the inherent difficulty of the pass due to factors such as distance and angle. Third, it lacks a mechanism to isolate the individual defender’s contribution from the joint effect of their teammates, making it difficult to disentangle shared defensive credit. This means that a defender can accumulate credit for low reception probabilities that are actually caused by external factors—such as the attacker being in an infeasible passing zone or the passer being under pressure—rather than the defender's own marking.

To simplify overload effects, one could restrict OBPR to Viterbi-assigned marking pairs rather than using ESO-weighted sums. However, this still does not address confounding influences from teammates and the game state. We also note that ESO is computed using the full observation sequence, which introduces potential information leakage from future events. For analyses concerned with causal interpretation, using forward probabilities (i.e., probabilities conditioned only on past observations) is a more appropriate approach.

\subsubsection*{Counterfactual Metrics Through CDHMM Generated Ghosts}
To better isolate individual defensive contributions, we outline a set of counterfactual metrics that compare the impact of a defender’s observed positioning, $d_{tj}$, to a baseline (ghost) position. These metrics quantify how a defender influences specific outcomes by measuring the difference in predicted values when the defender is placed at their actual location versus a simulated position generated from our CDHMM. We consider a \textit{Point-Ghost} baseline that substitutes the emission mean for the defender's position under a strict marking role. However, because our outcome prediction models are nonlinear in position, $\mathbb E[f(X)]\neq f(\mathbb E[X])$, point ghost comparisons introduce bias potentially underestimating the value of tight man-marking. If a defender should be shadowing on a particular side of the attacker then the tails of the ghost position distribution matter. We therefore adopt an \textit{Expected Ghost} baseline, integrating predictions over the role-conditioned emission through Monte Carlo sampling, and report that by default.
\[
\mathbb{E}_{x|k}\big[\,\cdot\,\big]\;:=\;\mathbb{E}_{x\sim p(d_{tj}\mid s_t=q_k)}\big[\,\cdot\,\big],
\]

While our empirical analysis focuses on pass reception probabilities, the same framework extends naturally to other outcome prediction models. In particular, we highlight how expected threat models could be used to capture a defender’s influence on downstream scoring potential, although we do not include empirical results using such models in this paper.

\begin{align}
\text{Reception Suppression:}\quad 
&\Delta^{(r)}_{tjk}= \mathbb{E}_{x|k}\!\left[ P_r(a_{tk}\mid x) \right] - P_r(a_{tk}\mid d_{tj}), \\
\text{Recovery Gain:}\quad 
&\Delta^{(\mathrm{rec})}_{tjk}= P_r(d_{tj}) - \mathbb{E}_{x|k}\!\left[ P_r(x) \right], \\
\text{Threat Suppression:}\quad 
&\Delta^{(\mathrm{th})}_{tjk}= \mathbb{E}_{x|k}\!\left[ P_{\mathrm{threat}}(a_{tk}\mid x) \right] - P_{\mathrm{threat}}(a_{tk}\mid d_{tj}), \\
\text{Counterattack Value:}\quad 
&\Delta^{(\mathrm{ca})}_{tjk}= P_{\mathrm{threat}}(d_{tj}) - \mathbb{E}_{x|k}\!\left[ P_{\mathrm{threat}}(x) \right].
\end{align}

\subsubsection*{Group-Level Evaluation}

While the above metrics evaluate individual attacker-defender interactions, they do not penalize defenders who fail to cover the most threatening attackers nearby. To address this, we introduce another metric, Group Coverage Advantage (GCA), which attempts to quantify if the defender's current position is advantageous compared to if they were strictly marking any of the attackers in a nearby region. In order to define which attackers we should consider we construct per-attacker attention weights using a softmax over ESO values:
\begin{equation}
w_{tjk} = \frac{\exp(\gamma_{tjk} / \tau)}{\sum_{\ell=1}^{J} \exp(\gamma_{t\ell k} / \tau)},
\end{equation}
where $\tau$ is a temperature parameter. This attacker-level vector shows which defenders could be reasonably affecting the attacker at each timestep, taking into account every defender's possible marking interactions with the attacker. We then define the set of attackers $K_{jt}$ that defender $j$ could plausibly influence at time $t$, 
\begin{equation}
K_{jt} = \left\{ k \;:\; w_{tjk} \geq \theta \right\},
\end{equation}
for threshold $\theta$, where tuning $\theta$ determines a minimum level of interaction between that defender and the attacker relative to the defender's teammates.

Let $f_{tk}(x)$ be a general function that predicts an outcome success probability for attacker $k$ given a defender's position $x$. We define the total group score as
\begin{equation}
G_{jt}(x) = \sum_{k \in K_{jt}} f_{tk}(x).
\end{equation}
If for example  $f_{tk}(d_{tj}) = P_r(a_{tk} \mid d_{tj})$, the total group score would represent the total reception probability of attackers that the defender could feasibly choose to mark given his own and teammates' positions. In reality, a defender's positioning will likely be a mixture of weighting themselves towards multiple opponents and passing avenues, in order to evaluate the defender's observed positioning we calculate $|K_{jt}|$ counterfactual baselines where we simulate the defender's position from the emission distribution corresponding to man-marking each attacker $k' \in K_{jt}$. The optimal baseline score over this group is then,
\begin{equation}
G^{\text{opt}}_{jt} = \min_{k' \in K_{jt}} \mathbb{E}_{x|k'}[G_{jt}(x)],
\end{equation}
the total outcome success probability that the defender would concede if they chose the most advantageous of all feasible man-marking roles, noting that the $\min$ operator may be replaced by $\max$ depending on the outcome model used (e.g., maximizing recovery probability instead of minimizing threat).

Finally, we define the Group Coverage Advantage at time $t$ as
\begin{equation}
\text{GCA}_j(t) = G^{\text{opt}}_{jt} - G_{jt}(d_{tj}),
\end{equation}
which measures how much additional value (e.g., threat) is allowed by the real defender’s positioning compared to the best of the ghost baselines over the group of feasible attackers.

\section*{Data availability}
The tracking and event datasets analysed in this study are subject to third-party licensing agreements and club confidentiality restrictions. As a result, the raw data cannot be shared publicly.

\section*{Code availability}
The code used for data processing, model training, and analysis was developed in a proprietary club context and is not publicly available. The authors are happy to answer specific implementation questions to support reproducibility of the methodology on independently obtained data.

\section*{Acknowledgments}
We thank the performance analysis staff at Nottingham Forest FC (NFFC) for their support and domain expertise. In particular, we acknowledge the analysts who produced the human baseline annotations used for evaluation of the corner reception prediction task. We also thank colleagues involved in establishing the industry–PhD programme that supported this work. This research was supported by NFFC, however the views expressed are those of the authors and do not necessarily reflect those of NFFC.

\section*{Author contributions}
SG led the study design, developed the methodology, implemented the models, curated the data, performed the experiments and analysis, and wrote the original manuscript.
SW supervised the research, contributed to study design and methodological decisions, and critically revised the manuscript.
FB provided project support and domain guidance, contributed feedback on modelling choices and interpretation, and critically revised the manuscript.
AR contributed to model design and study design through set-piece domain discussions, supported interpretation of results, and critically revised the manuscript.
LA contributed to supervision and provided feedback on interpretation and framing, and critically revised the manuscript.
\section*{Competing interests}
This research was conducted in collaboration with a professional football club that may have a competitive interest in the study outcomes. Due to confidentiality and competitive considerations, data and code are not publicly shared. The authors declare no other competing interests.

\bibliographystyle{unsrt}
\bibliography{references}

@article{RN4,
   author = {Forcher, Leander and Altmann, Stefan and Forcher, Leon and Jekauc, Darko and Kempe, Matthias},
   title = {The use of player tracking data to analyze defensive play in professional soccer - A scoping review},
   journal = {International journal of sports science and coaching},
   volume = {17},
   number = {6},
   pages = {1567-1592},
   ISSN = {1747-9541, 2048-397X},
   DOI = {10.1177/17479541221075734},
   url = {http://dx.doi.org/10.1177/17479541221075734},
   year = {2022},
   type = {Journal Article}
}

@article{Spatio-Temporal-Analysis-of-Team-Sports,
   author = {Gudmundsson, J. and Horton, M.},
   title = {Spatio-Temporal Analysis of Team Sports},
   journal = {Acm Computing Surveys},
   volume = {50},
   number = {2},
   note = {Gudmundsson, Joachim Horton, Michael
1557-7341},
   ISSN = {0360-0300},
   DOI = {10.1145/3054132},
   url = {<Go to ISI>://WOS:000405193800006},
   year = {2017},
   type = {Journal Article}
}

@article{Scofano_2024,
   title={About Latent Roles in Forecasting Players in Team Sports},
   volume={56},
   ISSN={1573-773X},
   url={http://dx.doi.org/10.1007/s11063-024-11532-0},
   DOI={10.1007/s11063-024-11532-0},
   number={2},
   journal={Neural Processing Letters},
   publisher={Springer Science and Business Media LLC},
   author={Scofano, Luca and Sampieri, Alessio and Re, Giuseppe and Almanza, Matteo and Panconesi, Alessandro and Galasso, Fabio},
   year={2024},
   month=feb }

@article{Fassmeyer2025Interactive,
  author    = {Fassmeyer, Daniel and Cordes, Martin and Brefeld, Ulrich},
  title     = {Interactive sequential generative models for team sports},
  journal   = {Machine Learning},
  volume    = {114},
  number    = {1},
  pages     = {38},
  year      = {2025},
  doi       = {10.1007/s10994-024-06648-2},
  url       = {https://doi.org/10.1007/s10994-024-06648-2}
}

@inproceedings{62f0eb3414504f6dae4f8e5f6621bd43,
title = "Modeling Defensive Dynamics in Football: A Hidden Markov Model-Based Approach for Man-Marking and Zonal Defending Corner Analysis",
author = "Sean Groom and Dan Morris and Liam Anderson and Shuo Wang",
note = "Not yet published as of 16/07/2024; The 2nd International Workshop on Intelligent Technologies for Precision Sports Science (IT4PSS) : in Conjunction with the 33rd International Joint Conference on Artificial Intelligence (IJCAI'24), IT4PSS ; Conference date: 04-08-2024",
year = "2024",
month = jun,
day = "5",
language = "English",
booktitle = "The 2nd International Workshop on Intelligent Technologies for Precision Sports Science (IT4PSS) in Conjunction with the 33rd International Joint Conference on Artificial Intelligence (IJCAI'24)",
url = "https://wasn.csie.ncu.edu.tw/workshop/IT4PSS2024.html",
}

@inproceedings{jain2019attention,
  title     = {Attention is Not Explanation},
  author    = {Jain, Sarthak and Wallace, Byron C.},
  booktitle = {Proceedings of the 2019 Conference of the North American Chapter of the Association for Computational Linguistics: Human Language Technologies (NAACL-HLT)},
  year      = {2019},
  pages     = {3543--3556},
  publisher = {Association for Computational Linguistics}
}

@inproceedings{wiegreffe2019attention,
  title     = {Attention is Not Not Explanation},
  author    = {Wiegreffe, Sarah and Pinter, Yuval},
  booktitle = {Proceedings of the 2019 Conference on Empirical Methods in Natural Language Processing (EMNLP-IJCNLP)},
  year      = {2019},
  pages     = {11--20},
  publisher = {Association for Computational Linguistics}
}

@article{match-analysis-big-data-and-tactics-current-trends-in-elite-soccer,
   author = {Memmert, D. and Rein, R.},
   title = {Match Analysis, Big Data and Tactics: Current Trends in Elite Soccer},
   journal = {Deutsche Zeitschrift f¸r Sportmedizin},
   volume = {69},
   number = {3},
   pages = {65-72},
   ISSN = {ISSN 0344-5925 (Print)
ISSN 2510-5264 (Electronic)},
   DOI = {10.5960/dzsm.2018.322},
   url = {https://www.germanjournalsportsmedicine.com/archive/archive-2018/heft-3/match-analysis-big-data-and-tactics-current-trends-in-elite-soccer/},
   year = {2018},
   type = {Journal Article}
}

@article{RN21,
   author = {Fernández, J. and Bornn, L. and Cervone, D.},
   title = {A framework for the fine-grained evaluation of the instantaneous expected value of soccer possessions},
   journal = {Mach Learn},
   volume = {110},
   number = {6},
   pages = {1389-1427},
   note = {1573-0565
Fernández, Javier
Orcid: 0000-0002-6297-042x
Bornn, Luke
Cervone, Daniel
Journal Article
United States
2021/11/12
Mach Learn. 2021;110(6):1389-1427. doi: 10.1007/s10994-021-05989-6. Epub 2021 May 24.},
   ISSN = {0885-6125 (Print)
0885-6125},
   DOI = {10.1007/s10994-021-05989-6},
   year = {2021},
   type = {Journal Article}
}

@inproceedings{RN22,
   author = {Decroos, T. and Bransen, L. and Van Haaren, J. and Davis, J. and Assoc Comp, Machinery},
   title = {Actions Speak Louder than Goals: Valuing Player Actions in Soccer},
   booktitle = {25th ACM SIGKDD International Conference on Knowledge Discovery and Data Mining (KDD)},
   pages = {1851-1861},
   note = {Decroos, Tom Bransen, Lotte Van Haaren, Jan Davis, Jesse
Davis, Jesse J/A-3596-2015
Bransen, Lotte/0000-0002-0612-7999; Davis, Jesse/0000-0002-3748-9263},
   ISBN = {978-1-4503-6201-6},
   DOI = {10.1145/3292500.3330758},
   url = {<Go to ISI>://WOS:000485562501090},
   year = {2019},
   type = {Conference Proceedings}
}

@article{RN1,
   author = {Rahimian, Pegah and Toka, Laszlo},
   title = {A data-driven approach to assist offensive and defensive players in optimal decision making},
   journal = {International Journal of Sports Science and Coaching},
   volume = {0},
   number = {0},
   pages = {17479541221149481},
   DOI = {10.1177/17479541221149481},
   url = {https://journals.sagepub.com/doi/abs/10.1177/17479541221149481
https://journals.sagepub.com/doi/10.1177/17479541221149481},
   type = {Journal Article}
}

@misc{veličković2018graphattentionnetworks,
      title={Graph Attention Networks}, 
      author={Petar Veličković and Guillem Cucurull and Arantxa Casanova and Adriana Romero and Pietro Liò and Yoshua Bengio},
      year={2018},
      eprint={1710.10903},
      archivePrefix={arXiv},
      primaryClass={stat.ML},
      url={https://arxiv.org/abs/1710.10903}, 
}

@misc{brody2022attentivegraphattentionnetworks,
      title={How Attentive are Graph Attention Networks?}, 
      author={Shaked Brody and Uri Alon and Eran Yahav},
      year={2022},
      eprint={2105.14491},
      archivePrefix={arXiv},
      primaryClass={cs.LG},
      url={https://arxiv.org/abs/2105.14491}, 
}

@book{Goodfellow2016DeepLearning,
  title={Deep Learning},
  author={Goodfellow, Ian and Bengio, Yoshua and Courville, Aaron},
  year={2016},
  publisher={MIT Press}
}

@article{rumelhart,
   author = {Rumelhart, David E. and Hinton, Geoffrey E. and Williams, Ronald J.},
   title = {Learning representations by back-propagating errors},
   journal = {Nature},
   volume = {323},
   number = {6088},
   pages = {533-536},
   ISSN = {1476-4687},
   DOI = {10.1038/323533a0},
   url = {https://doi.org/10.1038/323533a0},
   year = {1986},
   type = {Journal Article}
}

@inbook{ElementsOfInfomationTheory,
   author = {Thomas M. Cover, Joy A. Thomas},
   title = {Entropy, Relative Entropy, and Mutual Information},
   booktitle = {Elements of Information Theory},
   publisher = {Wiley},
   volume = {2nd Edition},
   pages = {13-55},
   DOI = {https://doi.org/10.1002/047174882X.ch2},
   url = {https://onlinelibrary.wiley.com/doi/abs/10.1002/047174882X.ch2},
   year = {2006},
   type = {Book Section}
}

@inproceedings{Le2017DataDrivenGUC,
  title={Data-Driven Ghosting using Deep Imitation Learning},
  author={Hoang Minh Le and Peter Carr and Yisong Yue and P. Lucey},
  booktitle={unknown},
  year={2017},
  url={http://www.yisongyue.com/publications/ssac2017\_ghosting.pdf}
}

@inproceedings{Seidl0BhostgustersRD,
  title={Bhostgusters : Realtime Interactive Play Sketching with Synthesized NBA Defenses},
  author={Thomas Seidl and A. Cherukumudi and Andrew Hartnett and Peter Carr and P. Lucey},
  booktitle={unknown},
  year={0},
  url={http://www.sloansportsconference.com/wp-content/uploads/2018/02/1006.pdf}
}

@article{RN30,
   author = {Higgins, Lewis and Galla, Tobias and Prestidge, Brian and Wyatt, Terry},
   title = {Measuring the pitch control of professional football players using spatiotemporal tracking data},
   journal = {Journal of Physics: Complexity},
   volume = {4},
   number = {2},
   pages = {025008},
   ISSN = {2632-072X},
   DOI = {10.1088/2632-072x/acb67d},
   url = {https://dx.doi.org/10.1088/2632-072x/acb67d},
   year = {2023},
   type = {Journal Article}
}

@inproceedings{RN31,
   author = {Taki, T. and Hasegawa, J. and Ieee Computer, Society and Ieee Computer, Society},
   title = {Visualization of dominant region in team games and its application to teamwork analysis},
   booktitle = {18th Computer Graphics International Conference of the Computer-Graphics-Society (CGI 2000)},
   pages = {227-235},
   note = {Taki, T Hasegawa, J},
   ISBN = {0-7695-0643-7},
   DOI = {10.1109/cgi.2000.852338},
   url = {<Go to ISI>://WOS:000088385100025},
   year = {2000},
   type = {Conference Proceedings}
}

@inproceedings{RN35,
   author = {Spearman, William and Basye, Austin Thomas and Dick, Gregory J. and Hotovy, Ryan and Hudl, Paul Pop},
   title = {Physics-Based Modeling of Pass Probabilities in Soccer},
   type = {Conference Proceedings}
}

@misc{kim2025defcon,
  title={Better Prevent than Tackle: Valuing Defense in Soccer Based on Graph Neural Networks},
  author={Kim, Hyunsung and Seo, Sangwoo and Choi, Hoyoung and Boomstra, Tom and Yoon, Jinsung and Park, Chanyoung},
  year={2025},
  eprint={2512.10355},
  archivePrefix={arXiv},
  primaryClass={cs.LG}
}

@inproceedings{everett2025gapp,
  title={Evaluating Defensive Influence in Multi-Agent Systems Using Graph Attention Networks},
  author={Everett, Gregory and Beal, Ryan J. and Matthews, Tim and Norman, Timothy J. and Ramchurn, Sarvapali D.},
  booktitle={2025 IEEE 12th International Conference on Data Science and Advanced Analytics (DSAA)},
  year={2025},
  doi={10.1109/DSAA65442.2025.11248003}
}

@article{RN8,
   author = {Wang, Zhe and Velickovic, Petar and Hennes, Daniel and Tomašev, Nenad and Prince, Laurel and Kaisers, Michael and Bachrach, Yoram and Élie, Romuald and Li, Wenliang Kevin and Piccinini, Federico and Spearman, William and Graham, Ian and Connor, Jerome T. and Yang, Yi and Recasens, Adrià and Khan, Mina and Beauguerlange, Nathalie and Sprechmann, Pablo and Moreno, Pol and Heess, Nicolas Manfred Otto and Bowling, Michael and Hassabis, Demis and Tuyls, Karl},
   title = {TacticAI: an AI assistant for football tactics},
   journal = {ArXiv},
   volume = {abs/2310.10553},
   year = {2023},
   type = {Journal Article}
}

@article{RN2,
   author = {Franks, A. and Miller, A. and Bornn, L. and Goldsberry, K.},
   title = {CHARACTERIZING THE SPATIAL STRUCTURE OF DEFENSIVE SKILL IN PROFESSIONAL BASKETBALL},
   journal = {Annals of Applied Statistics},
   volume = {9},
   number = {1},
   pages = {94-121},
   note = {Franks, Alexander Miller, Andrew Bornn, Luke Goldsberry, Kirk
Franks, Alexander/0000-0002-9329-206X},
   ISSN = {1932-6157},
   DOI = {10.1214/14-aoas799},
   url = {<Go to ISI>://WOS:000358354400005},
   year = {2015},
   type = {Journal Article}
}

@article{RN45,
   author = {Rabiner, L. R.},
   title = {A TUTORIAL ON HIDDEN MARKOV-MODELS AND SELECTED APPLICATIONS IN SPEECH RECOGNITION},
   journal = {Proceedings of the Ieee},
   volume = {77},
   number = {2},
   pages = {257-286},
   note = {Rabiner, lr
1558-2256},
   ISSN = {0018-9219},
   DOI = {10.1109/5.18626},
   url = {<Go to ISI>://WOS:A1989U374600002},
   year = {1989},
   type = {Journal Article}
}

@article{Merhej2021WhatHNA,
  title={What Happened Next? Using Deep Learning to Value Defensive Actions in Football Event-Data},
  author={Charbel Merhej and Ryan Beal and S. Ramchurn and Tim Matthews},
  journal={Proceedings of the 27th ACM SIGKDD Conference on Knowledge Discovery \& Data Mining},
  year={2021},
  url={https://api.semanticscholar.org/CorpusId:235313901}
}

@article{Forcher2023TheSFB,
  title={The Success Factors of Rest Defense in Soccer - A Mixed-Methods Approach of Expert Interviews, Tracking Data, and Machine Learning.},
  author={Leander Forcher and Leon Forcher and Stefan Altmann and D. Jekauc and M. Kempe},
  journal={Journal of sports science \& medicine},
  year={2023},
  volume={22 4},
  pages={
          707-725
        },
  url={https://api.semanticscholar.org/CorpusId:265071596}
}

@inproceedings{Yurko2024NFLGAA,
  title={NFL Ghosts: A framework for evaluating defender positioning with conditional density estimation},
  author={Ronald Yurko and Quang Nguyen and Konstantinos Pelechrinis},
  year={2024},
  url={https://api.semanticscholar.org/CorpusId:270711340}
}

@article{Tuyls2020GamePWB,
  title={Game Plan: What AI can do for Football, and What Football can do for AI},
  author={K. Tuyls and Shayegan Omidshafiei and Paul Muller and Zhe Wang and Jerome T. Connor and Daniel Hennes and I. Graham and W. Spearman and Tim Waskett and D. Steele and Pauline Luc and Adri{\`a} Recasens and Alexandre Galashov and Gregory Thornton and R. {\'E}lie and P. Sprechmann and Pol Moreno and Kris Cao and M. Garnelo and Praneet Dutta and Michal Valko and N. Heess and Alex Bridgland and J. P{\'e}rolat and B. D. Vylder and A. Eslami and Mark Rowland and Andrew Jaegle and R. Munos and T. Back and Razia Ahamed and Simon Bouton and Nathalie Beauguerlange and Jackson Broshear and T. Graepel and D. Hassabis},
  journal={ArXiv},
  year={2020},
  volume={abs/2011.09192},
  url={https://api.semanticscholar.org/CorpusId:227013043}
}

@inproceedings{Rudd2011MarkovSoccer,
  author    = {Sarah Rudd},
  title     = {A Framework for Tactical Analysis and Individual Offensive Production Assessment in Soccer Using Markov Chains},
  booktitle = {New England Symposium on Statistics in Sports (NESSIS)},
  address   = {Harvard University, Cambridge, MA},
  year      = {2011},
  month     = sep,
  note      = {Conference presentation; slides},
  url       = {https://nessis.org/nessis11/rudd.pdf},
  urldate   = {2025-09-16}
}

@misc{Singh2018xT,
  author       = {Karun Singh},
  title        = {Introducing Expected Threat (xT)},
  year         = {2018},
  month        = dec,
  howpublished = {\url{https://karun.in/blog/expected-threat.html}},
  note         = {Blog post},
  urldate      = {2025-09-16}
}

@article{Cervone2016EPV,
  author  = {Cervone, Daniel and D'Amour, Alex and Bornn, Luke and Goldsberry, Kirk},
  title   = {A Multiresolution Stochastic Process Model for Predicting Basketball Possession Outcomes},
  journal = {Journal of the American Statistical Association},
  year    = {2016},
  volume  = {111},
  number  = {514},
  pages   = {585--599},
  doi     = {10.1080/01621459.2016.1141685},
  url     = {https://doi.org/10.1080/01621459.2016.1141685}
}

@article{RN90,
   author = {Olthof, Sigrid and Davis, Jesse},
   title = {Perspectives on data analytics for gaining a competitive advantage in football: computational approaches to tactics},
   journal = {Science and Medicine in Football},
   pages = {1-13},
   note = {doi: 10.1080/24733938.2025.2533784},
   ISSN = {2473-3938},
   DOI = {10.1080/24733938.2025.2533784},
   url = {https://doi.org/10.1080/24733938.2025.2533784},
   type = {Journal Article}
}

@inproceedings{routineInspection,
   author = {Shaw, Laurie and Gopaladesikan, Sudarshan},
   title = {Routine inspection: A playbook for corner kicks},
   booktitle = {Machine Learning and Data Mining for Sports Analytics: 7th International Workshop, MLSA 2020, Co-located with ECML/PKDD 2020, Ghent, Belgium, September 14–18, 2020, Proceedings 7},
   publisher = {Springer},
   pages = {3-16},
   ISBN = {3030649113},
    year  = {2020},
   type = {Conference Proceedings}
}

@book{Bishop,
   author = {Bishop, Christopher M.},
   title = {Pattern recognition and machine learning},
   publisher = {New York : Springer, [2006] ©2006},
   note = {Textbook for graduates.;Includes bibliographical references (pages 711-728) and index.},
   url = {https://search.library.wisc.edu/catalog/9910032530902121},
   year = {2006},
   type = {Book}
}

@article{Kuhn,
   author = {Kuhn, H. W.},
   title = {The Hungarian method for the assignment problem},
   journal = {Naval Research Logistics Quarterly},
   volume = {2},
   number = {1-2},
   pages = {83-97},
   ISSN = {0028-1441},
   DOI = {https://doi.org/10.1002/nav.3800020109},
   url = {https://onlinelibrary.wiley.com/doi/abs/10.1002/nav.3800020109},
   year = {1955},
   type = {Journal Article}
}

@article{Munkres,
   author = {Munkres, James},
   title = {Algorithms for the Assignment and Transportation Problems},
   journal = {Journal of the Society for Industrial and Applied Mathematics},
   volume = {5},
   number = {1},
   pages = {32-38},
   ISSN = {03684245},
   url = {http://www.jstor.org/stable/2098689},
   year = {1957},
   type = {Journal Article}
}

@article{Bauer,
   author = {Bauer, Pascal and Anzer, Gabriel and Smith, Joshua Wyatt},
   title = {Individual role classification for players defending corners in football (soccer)},
   journal = {Journal of Quantitative Analysis in Sports},
   volume = {18},
   number = {2},
   pages = {147-160},
   DOI = {doi:10.1515/jqas-2022-0003},
   url = {https://doi.org/10.1515/jqas-2022-0003},
   year = {2022},
   type = {Journal Article}
}

\section*{Appendix}
\subsection*{Transition Model Dynamics}

\begin{figure}[htbp]
    \centering
\includegraphics[width=\linewidth]{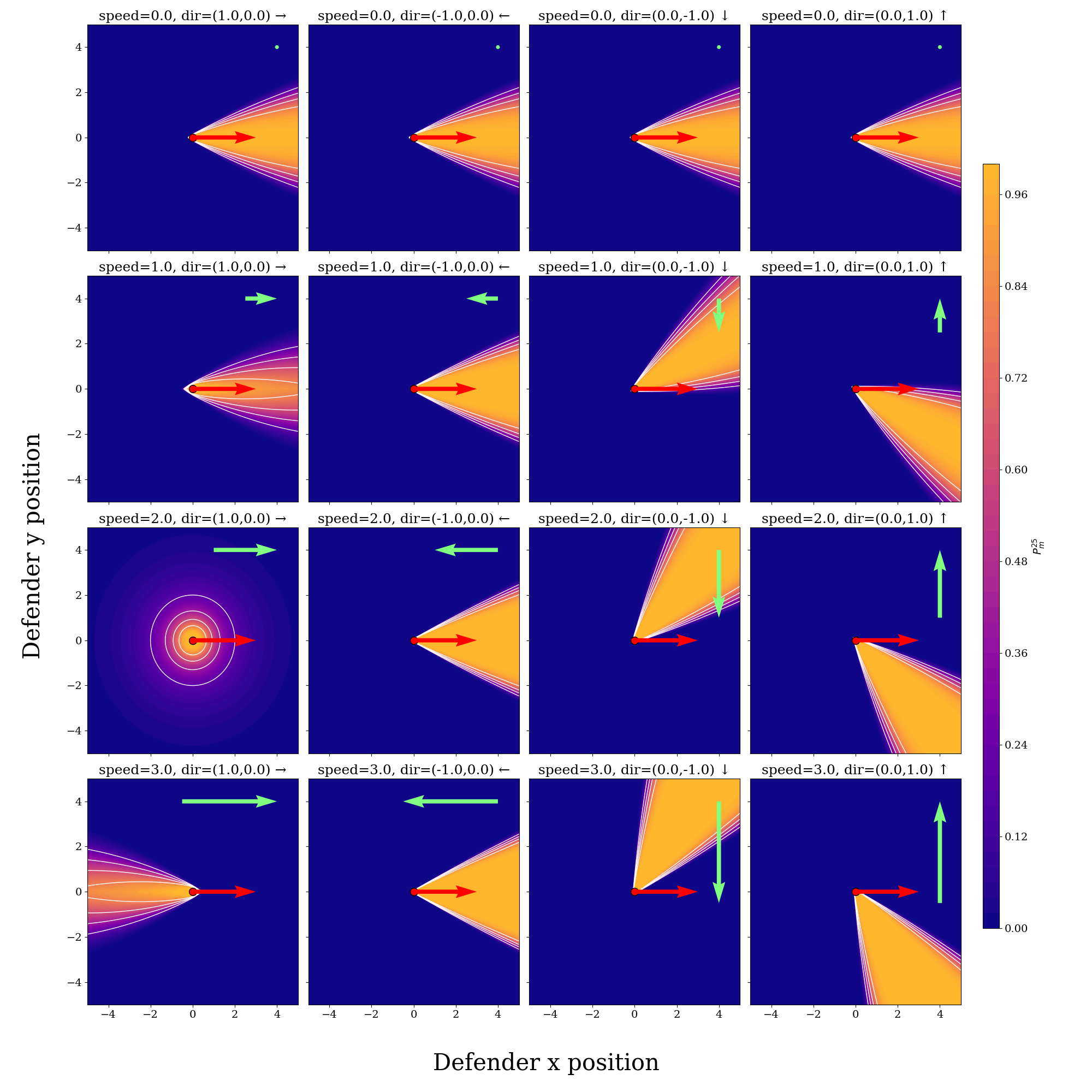}
    \caption{
        Each subplot shows how the probability of continuing to man-mark a specific attacker ($p_m$) varies spatially under different defender velocities. The attacker is fixed at the origin $(0, 0)$ and moves rightward with velocity $(2.0, 0.0)$, with height $1.85\,\text{m}$ and weight $82\,\text{kg}$. In each column, the defender moves in the same direction (indicated by the green arrow and subfigure title), with speed increasing from top to bottom. From left to right, each column shows the defender moving towards the left, right, down and up. Red arrows indicate the attacker's velocity vector. We plot three iso-probability contours at $p_m = 0.25$, $0.5$, and $0.75$ to highlight decision boundaries. Note that since data is sampled at 25 frames per second, the probability of maintaining a man-marking assignment is given by $p_m^{25}$.
        }

    \label{fig:transition heatmaps}
\end{figure}

Figure~\ref{fig:transition heatmaps} shows further examples of how the probability of continuing to mark a attacker for one second varies as distance and relative velocity of the man-marking pair varies.

\appendix
\subsection*{Sensitivity Analysis of the CDHMM}
\label{app:sensitivity}

Corner kicks are relatively infrequent and may exhibit tactical variation over the course of a season. For practical, in-season usage---where a team-specific model may be retrained week-by-week as new corners occur---it is therefore important to understand how the CDHMM behaves under varying amounts of training data. In this appendix, we report a sensitivity analysis that quantifies how learned parameters (zonal structure and transition weights) and overall fit (observation likelihood) change as additional sequences become available.

\subsubsection*{Experimental design}
\label{app:sensitivity_design}

For each team and delivery type (inswinging and outswinging), corner sequences from the 2023/24 season were ordered chronologically and incrementally grouped into training sets by adding batches of ten sequences. At each training set size, we trained ten CDHMM instances using different random seeds to account for variability due to EM initialization.

This procedure reflects realistic deployment, but it blends two sources of variation: (i) finite-sample estimation noise and (ii) potential distribution shift due to evolving execution or tactics over time. Rather than attempting to isolate these effects, we characterise their combined influence on model reliability under in-season retraining.

Unless stated otherwise, sensitivity results pool outcomes across all teams that had sufficient data to support training at a given sample size. Bar charts beneath each plot indicate the number of teams contributing at each training set size.

\subsubsection*{Stability metrics}
\label{app:sensitivity_metrics}

We evaluate sensitivity along three dimensions:

\begin{itemize}
    \item \textbf{Zonal structure disagreement.} To assess stability of learned zonal structure, we match zonal Gaussians across model instances and measure the resulting mismatch cost. Concretely, for each pair of models trained at the same sample size, we compute the optimal assignment between their learned zonal Gaussians using the Hungarian algorithm, using the 2-Wasserstein distance as the matching cost. This cost captures differences in both Gaussian means and covariances. We report the average total matching cost across all model pairs (lower is more similar).
    
    \item \textbf{Transition weight disagreement.} To quantify consistency in the learned transition model parameters, we compute the average pairwise $\ell_2$ distance between learned weight vectors for each transition type. We report disagreement separately for $\beta_m$ (man-marking continuation), $\beta_z$ (zonal continuation), and $\beta_s$ (switching).
    
    \item \textbf{Model fit (normalized observation log-likelihood).} To evaluate overall fit, we record the final observation log-likelihood achieved at convergence for each trained model and normalize it by the total number of timesteps across all training sequences (per-frame log-likelihood). We report distributions of these normalized values at each sample size.
\end{itemize}

\subsubsection*{Zonal structure sensitivity}
\label{app:sensitivity_zones}

\begin{figure}[htbp]
    \centering
    \begin{subfigure}{0.7\linewidth}
        \includegraphics[width=\linewidth]{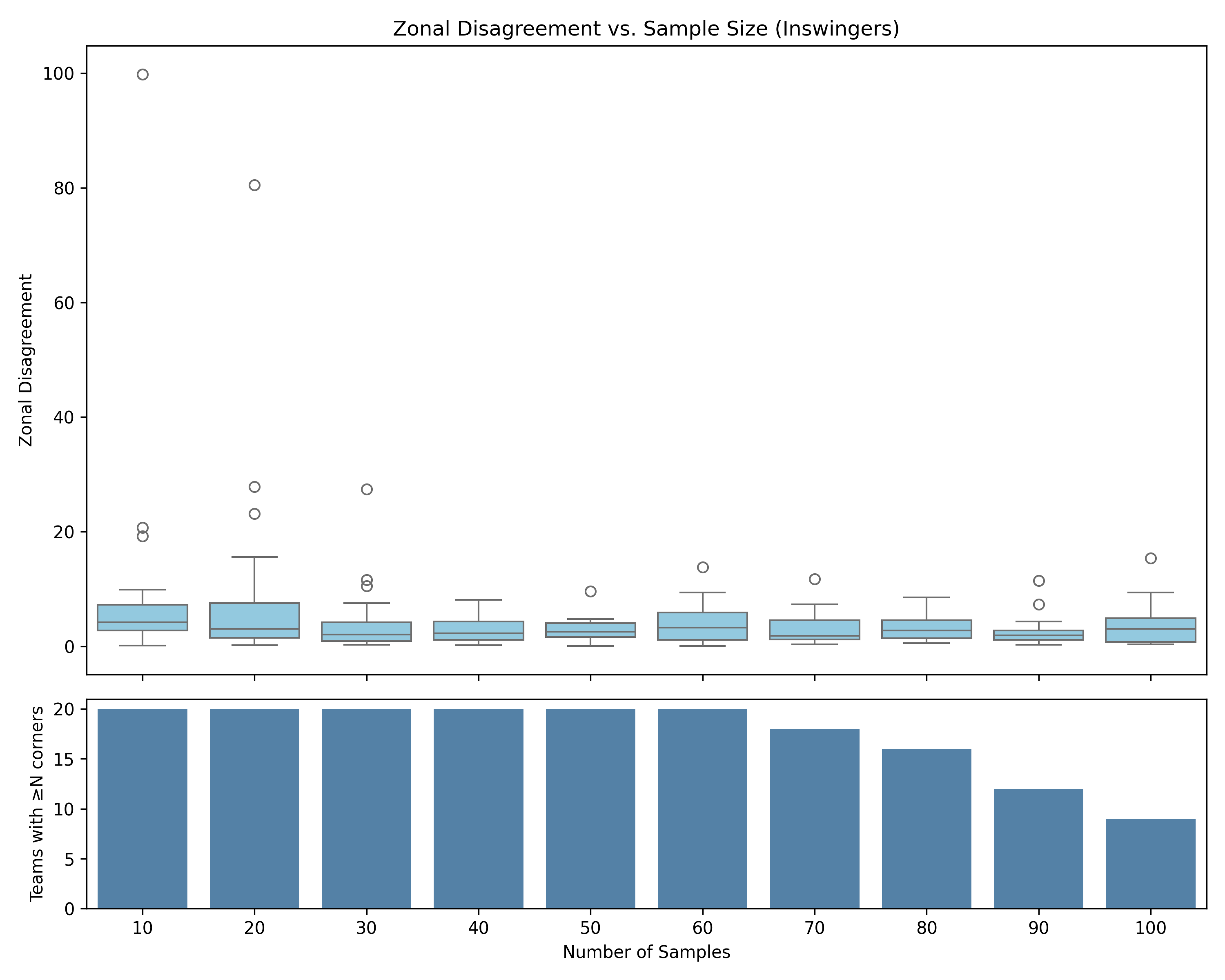}
        \caption{Inswing deliveries.}
        \label{fig:app_zonal_disagreement_inswing}
    \end{subfigure}

    \vspace{1em}

    \begin{subfigure}{0.7\linewidth}
        \includegraphics[width=\linewidth]{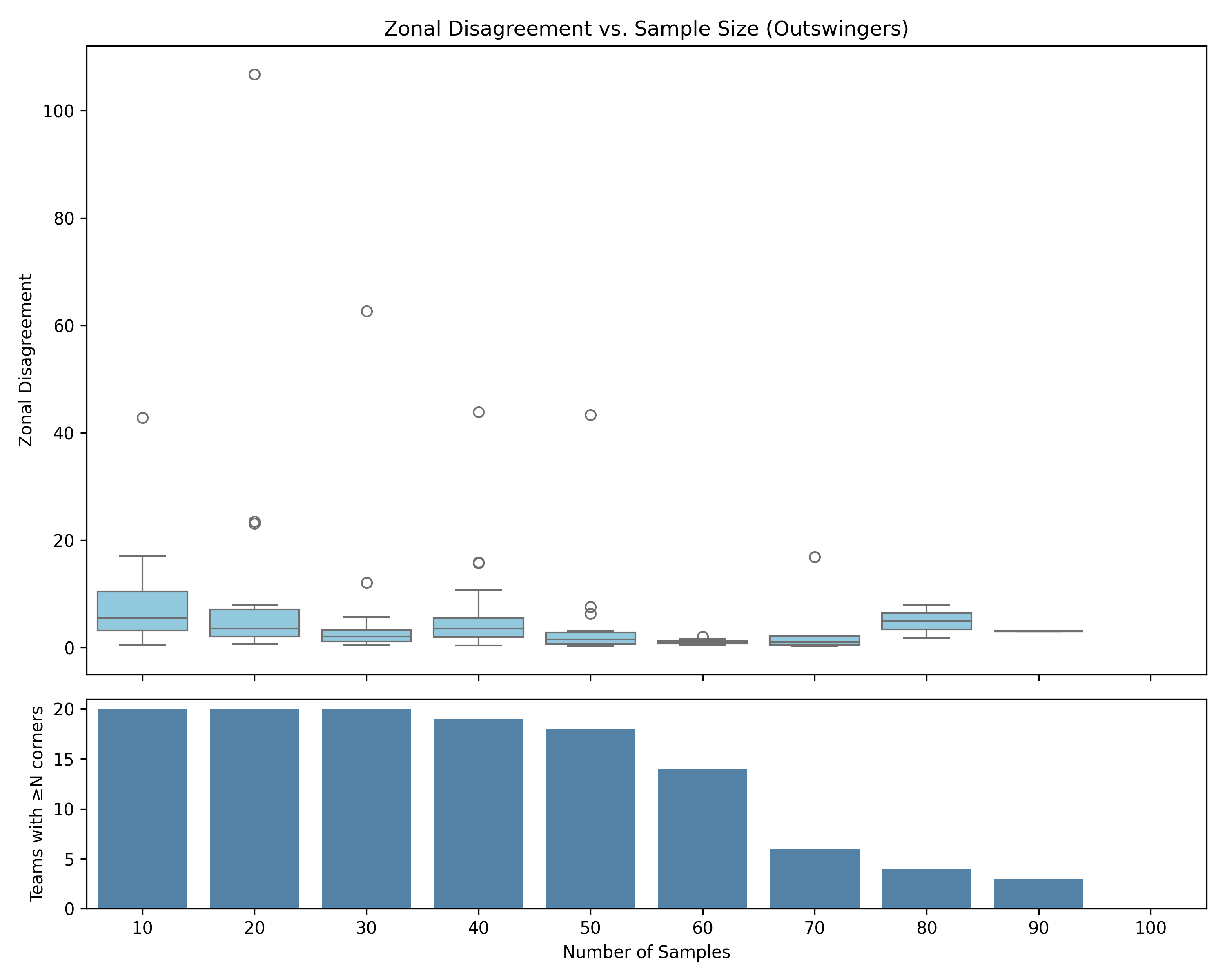}
        \caption{Outswing deliveries.}
        \label{fig:app_zonal_disagreement_outswing}
    \end{subfigure}

    \caption{Zonal structure disagreement as a function of training sample size for the CDHMM. Each box plot shows the distribution of pairwise zonal disagreement between models trained with the same sample size across all teams (10 random seeds per team), with outliers plotted. Bar plots indicate the number of teams contributing at each sample size.}
    \label{fig:app_zonal_disagreement}
\end{figure}

Figure~\ref{fig:app_zonal_disagreement} shows the distribution of average pairwise zonal structure disagreement across the 10 CDHMMs trained at each sample size, shown separately for inswinging and outswinging deliveries. Across both delivery types, zonal disagreement generally decreases as sample size increases, with both the median and interquartile range narrowing. This is consistent with reduced estimation noise as more data becomes available.

Despite the overall stabilisation trend, occasional large outliers persist even at higher sample sizes. These outliers do not consistently correspond to particular teams, suggesting they are not due to persistent modelling failures for specific defensive styles. Instead, they likely reflect corner-to-corner variability in execution or tactical structure: when a chronologically defined subset contains atypical sequences, the model can converge to a noticeably different zonal configuration. Practically, this indicates that while the CDHMM becomes more stable with additional data, a minority of retraining windows can yield substantially different zone estimates, motivating uncertainty-aware model monitoring when deployed across a season.

\subsubsection*{Transition weight sensitivity}
\label{app:sensitivity_transitions}

\begin{figure}[htbp]
    \centering
    \begin{subfigure}{0.7\linewidth}
        \includegraphics[width=\linewidth]{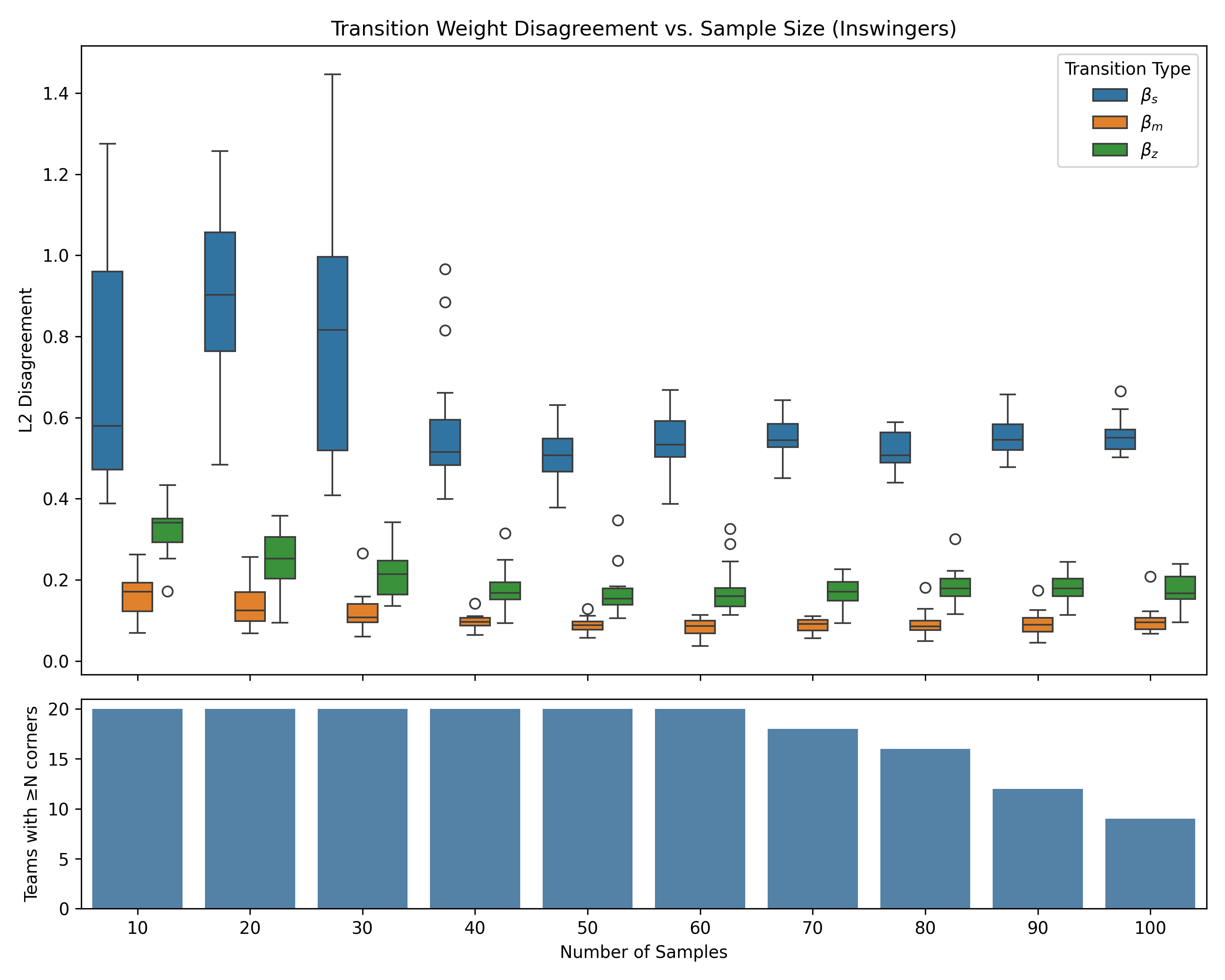}
        \caption{Inswing deliveries.}
        \label{fig:app_beta_disagreement_inswing}
    \end{subfigure}

    \vspace{1em}

    \begin{subfigure}{0.7\linewidth}
        \includegraphics[width=\linewidth]{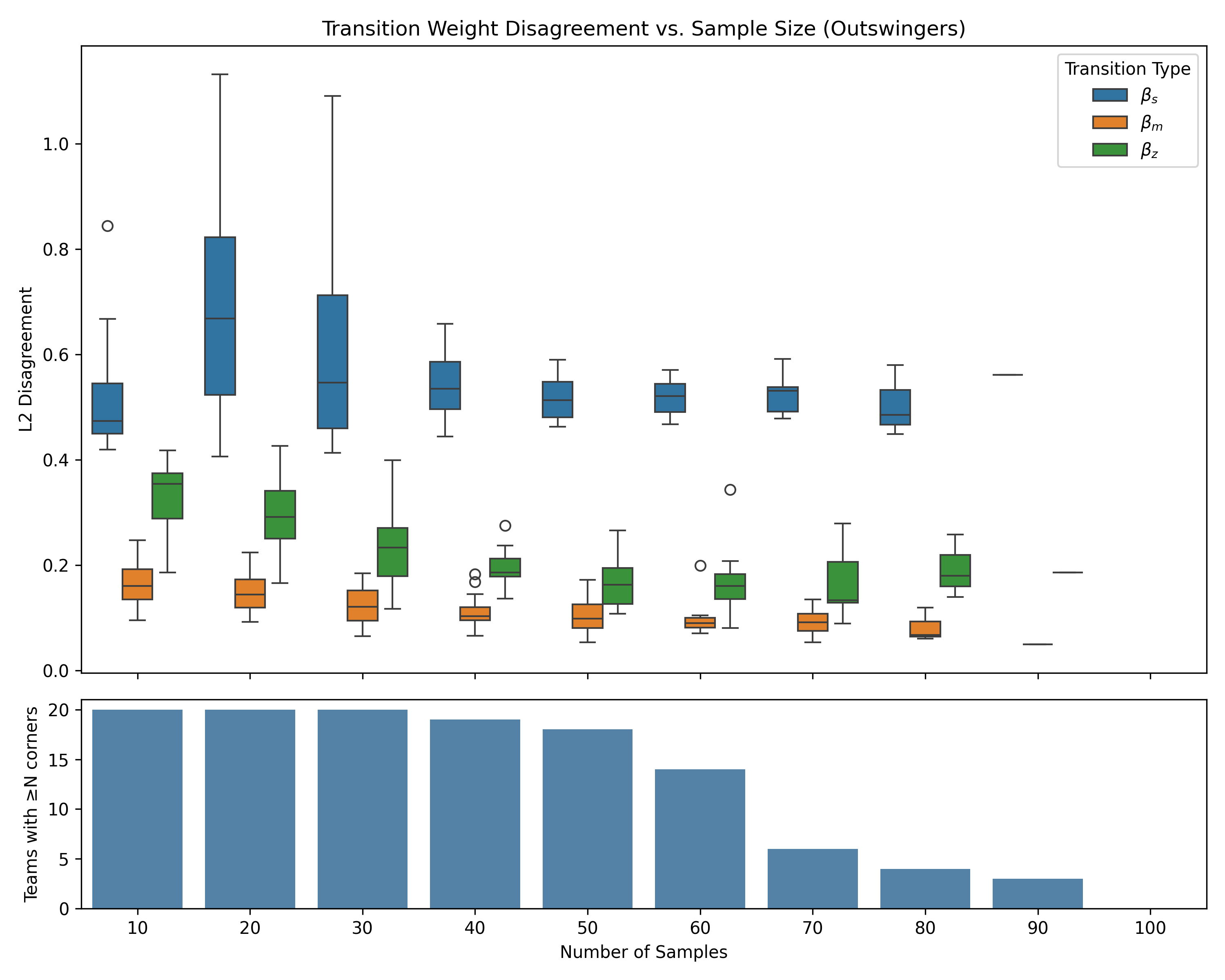}
        \caption{Outswing deliveries.}
        \label{fig:app_beta_disagreement_outswing}
    \end{subfigure}

    \caption{Disagreement in transition weight parameters as a function of training sample size for the CDHMM. Each box plot shows the distribution of pairwise $\ell_2$ distances between transition weights ($\beta_s$, $\beta_m$, $\beta_z$) learned from models trained on the same sample size (10 random seeds per team), pooled across all teams with sufficient data. Outliers are plotted. Bar plots indicate the number of teams contributing at each sample size.}
    \label{fig:app_beta_disagreement}
\end{figure}

Figure~\ref{fig:app_beta_disagreement} reports disagreement in transition weights as a function of training sample size. Two consistent patterns emerge. First, estimation behaviour is similar between inswinging and outswinging deliveries, yielding near-identical disagreement distributions for each parameter type. Second, the continuation weights $\beta_m$ and $\beta_z$ converge relatively quickly: pairwise $\ell_2$ disagreements drop and stabilise once modest numbers of corner samples are available. In contrast, the switching weights $\beta_s$ remain substantially more variable, likely because switch events are comparatively rare. This suggests that uncertainty in $\beta_s$ may persist even with additional data, motivating future work that shares statistical strength across teams (e.g., hierarchical priors or grouped models) when estimating infrequent transitions.

\subsubsection*{Normalized likelihood sensitivity}
\label{app:sensitivity_likelihood}

\begin{figure}[htbp]
    \centering
    \begin{subfigure}{0.7\linewidth}
        \includegraphics[width=\linewidth]{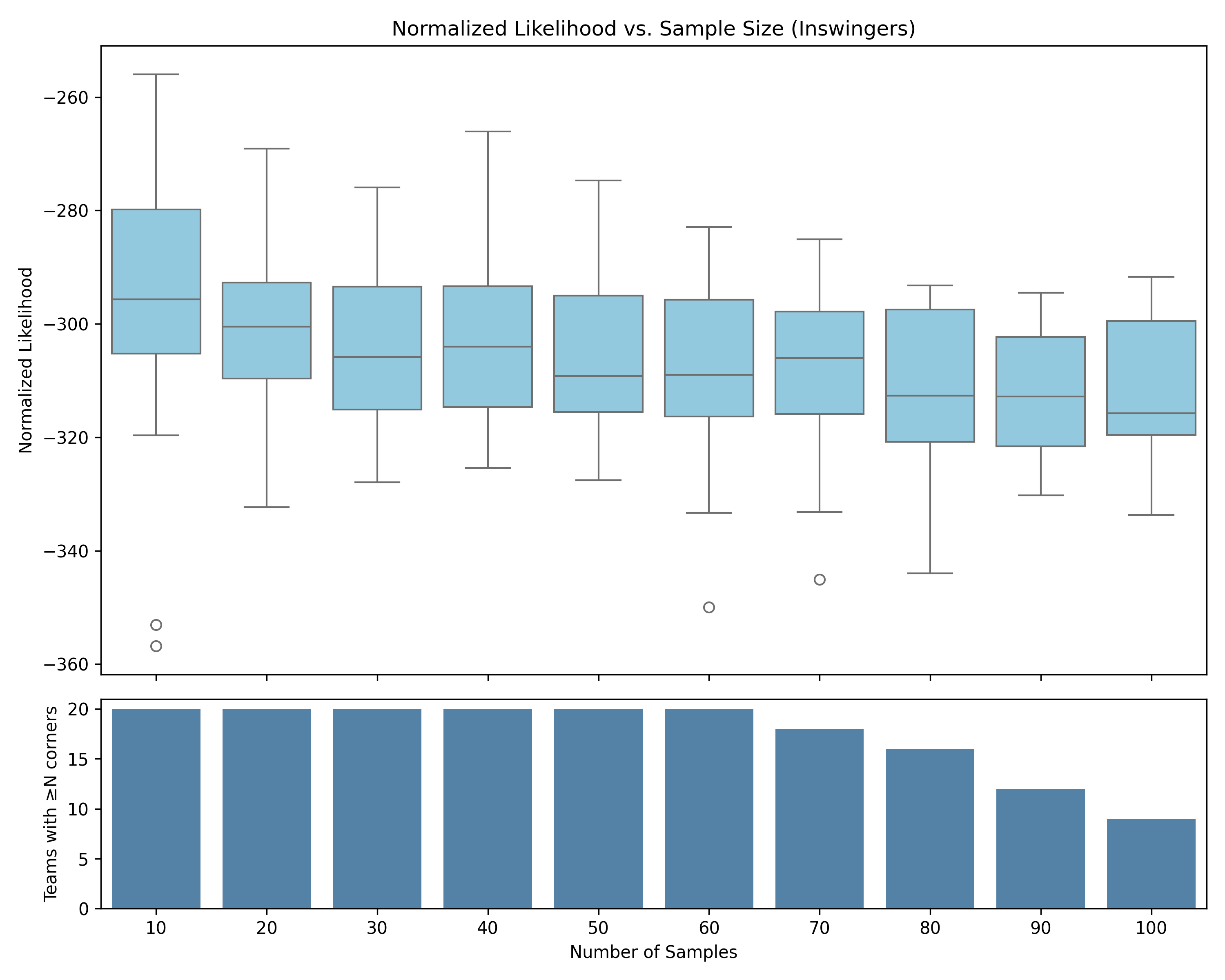}
        \caption{Inswing deliveries.}
        \label{fig:app_likelihood_inswing}
    \end{subfigure}

    \vspace{1em}

    \begin{subfigure}{0.7\linewidth}
        \includegraphics[width=\linewidth]{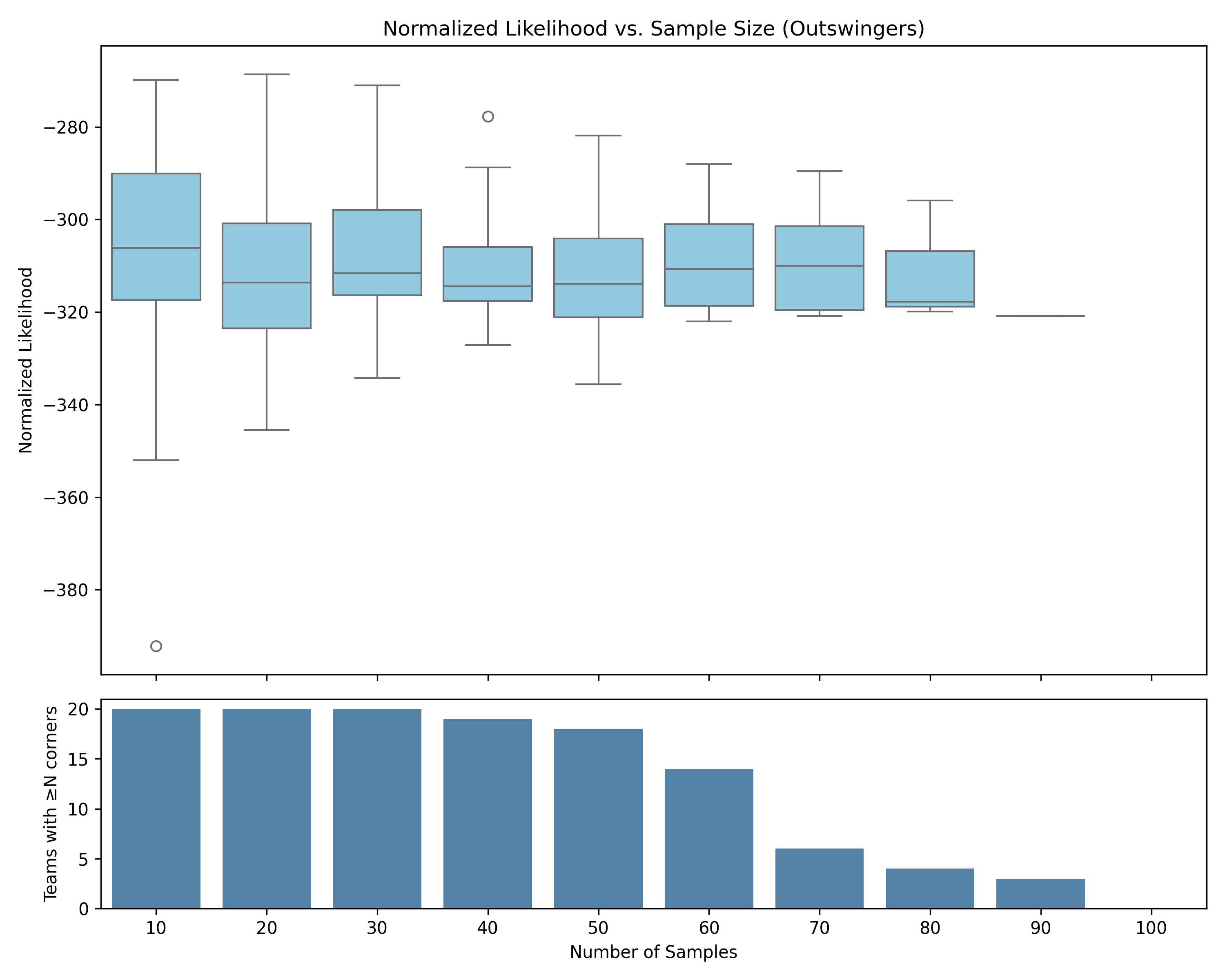}
        \caption{Outswing deliveries.}
        \label{fig:app_likelihood_outswing}
    \end{subfigure}

    \caption{Normalized observation log-likelihood as a function of training sample size for inswinging and outswinging deliveries. Each box plot shows the distribution of per-frame log-likelihoods from CDHMMs trained with different random seeds at the same sample size, pooled across teams with sufficient data. Log-likelihoods are normalized by the total number of timesteps across all training sequences. Bar plots indicate the number of teams contributing at each sample size.}
    \label{fig:app_likelihood}
\end{figure}

Figure~\ref{fig:app_likelihood} presents distributions of normalized observation log-likelihoods as training sample size increases. For both inswingers and outswingers, the average per-frame log-likelihood tends to become more negative as additional sequences are incorporated before approximately plateauing. Although the CDHMM is trained to maximise likelihood on the available dataset, this pattern is consistent with reduced overfitting as training sets expand: models trained on limited data can fit repeated or idiosyncratic behaviours unusually well, whereas larger datasets expose more tactical diversity and reduce the attainable average likelihood.

The spread of normalized likelihoods typically narrows at larger sample sizes, indicating improved consistency across random initializations and convergence toward a more stable representation of team behaviour. Inswinging deliveries achieve consistently higher normalized likelihoods than outswinging deliveries, which may indicate that inswing defending exhibits more structured or predictable patterns that are better aligned with the model assumptions.

\subsubsection*{Illustrative convergence of learned zones for a single team}
\label{app:sensitivity_overlay}

\begin{figure}[htbp]
    \centering
    \includegraphics[width=\linewidth]{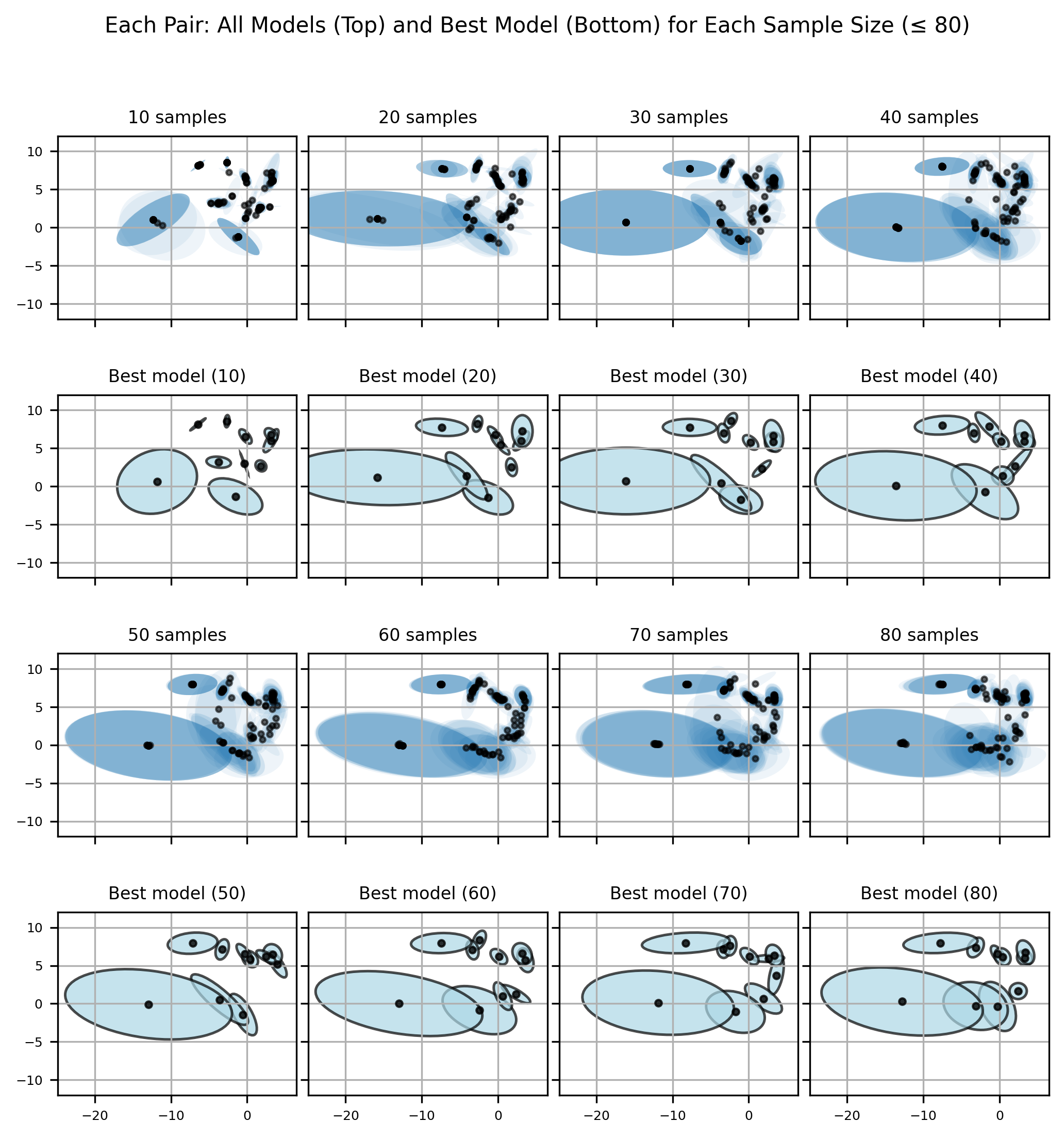}
    \caption{Stability of learned defensive zone positions across increasing training sample sizes for an illustrative team. The top row shows overlays of all 10 models trained at each sample size (each ellipse represents a bivariate Gaussian zone; means are marked as black dots), while the bottom row shows the single best model selected by maximum observation likelihood. As sample size increases, estimated zone positions become more stable and less variable across random initializations, while remaining differences across chronological subsets suggest minor tactical or execution changes over time.}
    \label{fig:app_zone_stability_overlay}
\end{figure}

To complement the aggregate disagreement metrics, Figure~\ref{fig:app_zone_stability_overlay} provides a qualitative example of how an individual team's zonal configuration converges as more corners are added. The top row overlays zonal Gaussians from all ten random initializations at each sample size, while the bottom row shows the single best model selected by maximum observation likelihood. As sample size increases, both within-sample variability (across random seeds) and between-sample variability (across training windows) diminish, indicating increasingly stable zone estimates. Residual variation in the best models suggests that chronological subsets can still differ due to changes in execution or tactics over time, consistent with the outliers observed in Figure~\ref{fig:app_zonal_disagreement}.

\end{document}